\documentclass{article}

 \usepackage[main, preprint]{neurips_2026}

\usepackage[utf8]{inputenc} \usepackage[T1]{fontenc}    \usepackage{hyperref}       \usepackage{url}            \usepackage{booktabs}       \usepackage{nicefrac}       \usepackage{microtype}      \usepackage{xcolor}         \usepackage{times}
\usepackage{latexsym}
\usepackage{xspace}
\usepackage{listings}

\usepackage{microtype}

\usepackage{inconsolata}

\usepackage{graphicx}

\usepackage{epigraph}
\usepackage{CJKutf8}

\usepackage{booktabs} 

\usepackage{algorithm}
\usepackage{algpseudocode}

\usepackage{xcolor}
\usepackage{enumitem} 
\usepackage[table]{xcolor}
\definecolor{oursrow}{gray}{0.94}
\usepackage{multirow}
\usepackage{amsmath,amsfonts,amsthm,bm}

\algrenewcommand{\algorithmiccomment}[1]{\hskip1em$\triangleright$ #1}

\newcommand{\method}{\texttt{SpeedRunner}\xspace}

\title{Towards Embodied Skill Learning\\via Online Program Induction}
\title{Embodied Symbolic Skill Learning\\via Online Program Induction}
\title{Embodied Skill Learning\\via Online Program Induction}
\title{You Can't Unscramble an Egg: Learning Programmatic Skills Without Experience Replay in Embodied Environments}
\title{You Can't Unscramble an Egg: Inducing Programmatic Skills for Lifelong Learning}
\title{Online Learning of Programmatic Skills}
\title{Programmatic Skill Learning Best Reduces Agent Cost}
\title{\method: Programmatic Skill Learning\\Best Reduces Agent Cost}
\title{Better, Faster, Stronger: Programmatic Skill\\Learning Best Reduces Agent Cost}

\author{Zixi Huang$^{*\dagger}$ \quad Xiheng Wang$^*$ \quad
Andrew Wang$^{*\dagger}$ \AND
William Jurayj \quad
Bernal Jiménez Gutiérrez \quad
Daniel Khashabi$^\dagger$ \quad
Nicholas Andrews$^\dagger$ 
\\
\\
Johns Hopkins University \quad \\
\texttt\{zhuang60, awang116, danielk, noa\}@jhu.edu \\
$^*$Equal contribution \quad $^{\dagger}$Corresponding authors
}

\begin{document}

\maketitle

\newcommand{\daniel}[1]{{\color{red} [Daniel: #1]}}

\definecolor{actorcolor}{RGB}{0,114,178}      \definecolor{inducercolor}{RGB}{126,61,104}   \definecolor{librarycolor}{RGB}{38,112,97}    \definecolor{historycolor}{RGB}{166,91,24}    
\newcommand{\actorvar}[1]{\textcolor{actorcolor}{#1}}
\newcommand{\inducervar}[1]{\textcolor{inducercolor}{#1}}
\newcommand{\libraryvar}[1]{\textcolor{librarycolor}{#1}}
\newcommand{\historyvar}[1]{\textcolor{historycolor}{#1}}

\newcommand{\actortext}[1]{\textcolor{actorcolor}{\textbf{#1}}}
\newcommand{\inducertext}[1]{\textcolor{inducercolor}{\textbf{#1}}}
\newcommand{\librarytext}[1]{\textcolor{librarycolor}{\textbf{#1}}}
\newcommand{\historytext}[1]{\textcolor{historycolor}{\textbf{#1}}}

\begin{abstract}

Recently, the practice of augmenting LLM agent capability with \emph{skills} has gained prevalence.
We explore the cost effective adaptation of agents to novel domains by means of learning skills.
Existing works focus on performance gain over cost effectiveness.
As a result, little is known about what skill learning strategies save cost.
We argue that among all the different skill learning methods, those that view skills as programs can achieve the best cost reduction.
By executing sequences of actions deterministically, a program-augmented agent can reliably and cheaply achieve goals that would otherwise require trial and error and risk degenerate behavior over long horizons.
An agent can learn at inference time by incrementally discovering these programs and equipping them for future tasks.
We hypothesize that past trajectories contain enough signal to guide skill learning, even without replay or validation, provided the agent can learn to analyze them.
To test our claims, we propose \method, a coding agent that analyzes trajectories and refactors skills for better performance on future tasks.
Across three different embodied environments, we show that \method consistently achieves the frontier in learning and cost reduction while remaining robust against distribution shifts and environmental randomness.

\end{abstract}

\section{Introduction}

\begin{figure*}[ht]
    \centering
    \includegraphics[width=\textwidth]{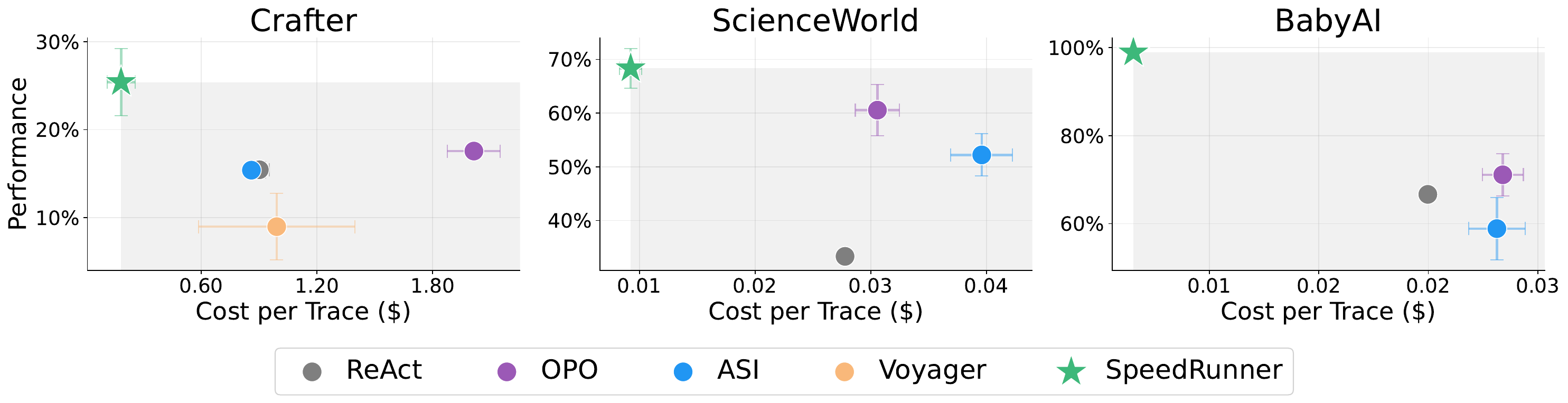}
    \caption{Performance vs average cost per trace for different approaches with GPT 5.4-mini on 3 benchmarks using learned programmatic skills. \textbf{\method consistently dominates baselines in cost and performance, highlighting the potential advantages of programmatic skills}.  
            }
        \label{fig:combined_performance_pareto}
\end{figure*}

As LLM agents become more prevalent, they are increasingly deployed in complex real world environments.
For instance, such agents now run a real world coffee shop in Stockholm.\footnote{\url{https://andonlabs.com/blog/ai-cafe-stockholm}}
Such deployments rarely present a fixed task distribution: new menu items and unfamiliar requests accumulate over time, so a static policy degrades unless the agent can adapt on the job.
In these settings, we expect agents to learn \emph{online} from their environment without catastrophically forgetting their pre-existing knowledge.
We view \emph{skill learning} as a promising paradigm. 
Here an agent reflects on its trajectories and updates its policy with environment-specific capabilities called \emph{skills} (\S\ref{sec:prelims}).
The manner in which the policy is updated, the representation of skills, and the degree of online vs offline learning, vary between methods.

The predominant representation of skills are natural language descriptions \citep{zhong2026skilllearnbenchbenchmarkingcontinuallearning, mi2026skillpro, ni2026trace2skilldistilltrajectorylocallessons, wang2026skillsdskillconditionedselfdistillationmultiturn, yang2026autoskillexperiencedrivenlifelonglearning}, following the release of Anthropic's agent skills .
However, this shift has overlooked a key consideration for the average user: \textbf{cost}.
Existing works focus on performance gains, with cost savings as an afterthought.
Thus the effect of different skill representation and learning strategies on cost remains unknown. 
\textbf{We argue that the modality of code can be a more cost effective way to represent skills, compared to natural language.}
Rather than reasoning about the same routine over and over again, an agent learns to offload that reasoning to a cheaper Turing machine. 
Hence, the more reasoning successfully translated into code, the more cost saved.

However, the key question is how well agents can learn programmatic skills from experience, and then successfully use them to accomplish tasks. 
Existing approaches fall short in multiple ways.
First, the prevailing method of learning from experience has been to prompt an LLM with past trajectories. 
This method is critically bottlenecked by trajectory length; the number that fit in context decreases as they grow longer, until none fit at all.
Second, unrealistic assumptions are made to facilitate learning, the most glaring of which assumes the ability to manipulate the environment to the extent of undoing and rewinding states.
This often manifests as a set of held-out episodes used to gather heuristics for hillclimbing that are replayed repeatedly.
Together, these shortcomings prevent the practical deployment of programmatic skill learning, especially in multi-turn agentic settings. 

We hypothesize that the agent trajectories themselves are signal rich, and that simply analyzing them properly resolves these problems.
\textbf{At a high level, we frame the process of extracting signal as an agentic coding task.}
Given a history of trajectories, an agent with access to a code execution environment can programmatically identify recurring behavior patterns, skill usage and success metrics, performance regression, among others.
Unlike natural language feedback \citep{agrawal2026gepareflectivepromptevolution,pryzant2023automaticpromptoptimizationgradient} or handcrafted tools \citep{stengel-eskin_regal_2024}, a coding agent can avoid needle-in-a-haystack retrieval over long trace histories, support more structured cross-trajectory analysis, and mitigate the context-window bottleneck that arises in long-horizon environments with 200-step solutions \citep{crafter}.

To test our hypothesis, we propose \method, a simple yet effective way to induce programmatic skills using coding agents to analyze observed trajectories.
Our goal is to optimize a stochastic policy parameterized by a library of skills, which, in practice, is an LLM agent equipped with a set of callable functions.
The search for optimal skills cycles between two phases.
The stochastic policy receives tasks and generates trajectories.
Then, based on the trajectories, \method updates the skills.

We evaluate \method on ScienceWorld \citep{wang-etal-2022-scienceworld}, Crafter \citep{crafter}, and BabyAI \citep{chevalier-boisvert2018babyai}---a diverse set of embodied simulations.
Each domain features vastly different dynamics, physical constraints, and action spaces.
Nevertheless, we find consistent trends across all domains.
(1) Programmatic skills can be learned online, without replay buffers or continual validation.
(2) Their primary advantage is cost reduction without compromising performance.
(3) Programmatic trajectory analysis enables skill induction for long-horizon agents.
We also find that programmatic skill learning can be robust to environmental randomness and to distribution shifts.

\section{Preliminaries} \label{sec:prelims}

\textbf{Problem formulation.} 
We assume a partially observed Markov decision process (POMDP) and a set of valid initial states, where task \emph{instructions} if available are part of the initial state. 
The POMDP we refer to as the \emph{environment} and the set of valid initial states as \emph{tasks} ($\mathcal{T}$).
Tasks are sampled one by one for a stochastic policy $\pi$ to complete.
A rollout results in a sequence of observations and actions ending in task reward $(o_1, a_1, o_2, a_2, ..., o_T, a_T, R)$.
We refer to this tuple as a \emph{trajectory} ($\tau$).

\textbf{Pure Online-ness.} In order to accommodate real world settings, we aim to require as little as possible from the environment.
Therefore, we do not assume the ability to undo and redo environment actions, as many settings such as customer service or high frequency trading preclude such possibility.
This decision leaves us with ``purely-online'' learning, where tasks are ephemeral and cannot be played again with an updated policy. 
All approaches and baselines we evaluate are made to conform to this standard.

\textbf{Defining skills.} 
Prior works have treated skills as prompt strategies \citep{wang_agent_2025}, executable functions \citep{stengel-eskin_regal_2024,wang2025inducingprogrammaticskillsagentic,zheng2025skillweaverwebagentsselfimprove}, or a mixture of both.\footnote{\url{https://platform.claude.com/docs/en/agents-and-tools/agent-skills/overview}}
We focus on executable functions because code is expressive, precise, reliably executed, and structurally analyzable through call graphs and abstract syntax trees.
We define a \emph{skill} as an executable function paired with documentation, operating on the atomic actions defined by the POMDP or other existing skills.
We refer to \emph{skill} when we mean a function and its documentation, and \emph{function} when we mean the code alone.

\section{\method: Effective Trajectory Analysis With Coding Agents}

\begin{figure*}
    \centering
    \includegraphics[scale=0.53,trim=1.2cm 8cm .8cm 4.3cm,clip=true]{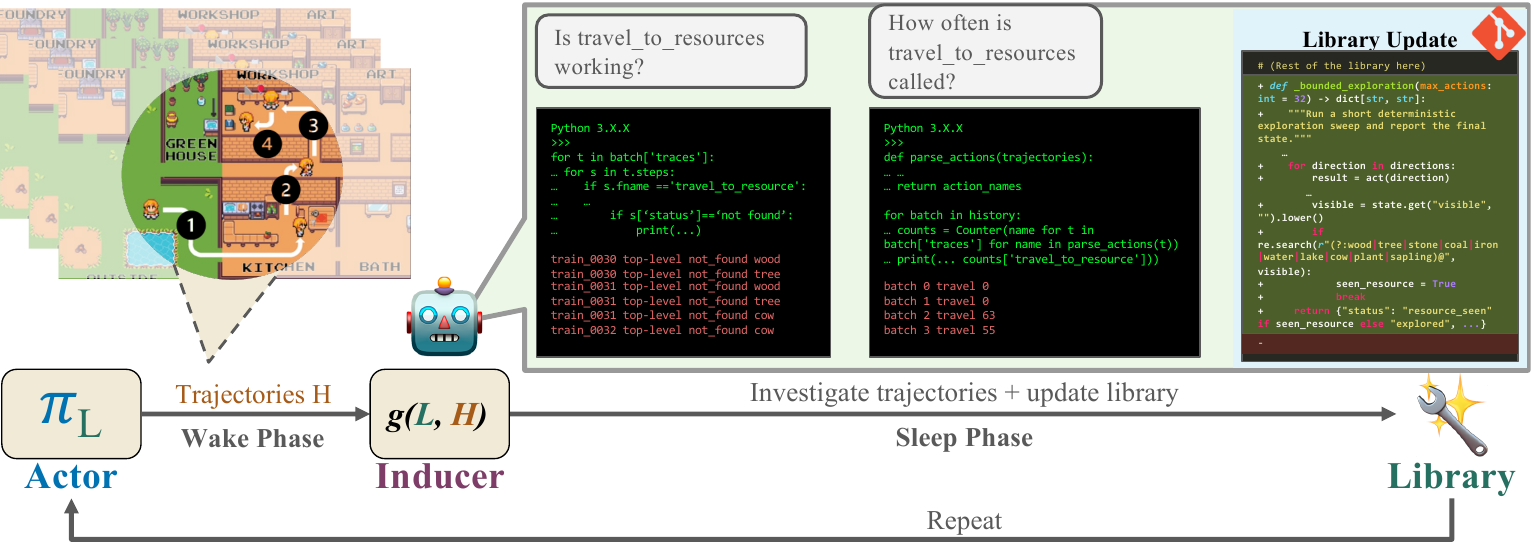}
    \caption{
Overview of \method.
A stochastic policy
$\actorvar{\pi}_{\libraryvar{L}}$ (the actor) is equipped with a
skill library $\libraryvar{L}$, while a proposal function
$\inducervar{g}$ (the inducer) updates the library using trajectory
history $\historyvar{H}$.
During the wake phase, the actor generates a batch of trajectories.
During the sleep phase, the inducer examines the trajectories and
edits the library.
In this example, the inducer notices repeated ``not found'' responses
from \texttt{travel\_to\_resource}, checks the function's usage
frequency, and introduces \texttt{\_bounded\_exploration}, which
searches the actor's local neighborhood for the desired resource.
}
    \label{fig:method-overview}
\end{figure*}
We parameterize a stochastic policy by a
\librarytext{skill library} $\libraryvar{L}$, writing the resulting
policy as $\actorvar{\pi}_{\libraryvar{L}}$.
Learning alternates between two phases.
During the \emph{wake} phase, the
\actortext{actor} $\actorvar{\pi}_{\libraryvar{L}}$ generates a batch
of trajectories $B$, which is appended to the
\historytext{history} $\historyvar{H}$.
During the \emph{sleep} phase, the
\inducertext{inducer}
$\inducervar{g}:
\libraryvar{L}\times\historyvar{H}\rightarrow\libraryvar{L}$
edits the library.
In our experiments, the actor is an LLM agent equipped with callable
skills, and the inducer is a coding agent.

\method focuses on the inducer: how should
$\inducervar{g}$ extract learning signal from
$\historyvar{H}$?
Trajectory-level rewards are available but too sparse to reliably guide library edits from only a few online episodes.
Richer signal lies inside trajectories as \emph{environment responses}: errors, unexpected outcomes, and other feedback from the actor's actions.
These signals are unstructured and often buried in long-horizon traces, making full-context reading expensive and summarization lossy.

\begin{algorithm}[t]
\caption{\method: Wake--Sleep Skill Library Learning}
\label{alg:speedrunner}
\begin{algorithmic}[1]
\Require initial library $\libraryvar{L_0}$,
actor $\actorvar{\pi}_{\libraryvar{L}}$,
inducer
$\inducervar{g}:
\libraryvar{L}\times\historyvar{H}\to\libraryvar{L}$,
environment $E$, task distribution $\mathcal{T}$,
total episodes $N$, minibatch size $k$

\State $\historyvar{H} \gets \emptyset$
\State $\libraryvar{L} \gets \libraryvar{L_0}$

\For{$n = 1, \dots, N$ \textbf{by} $k$}
    \State \Comment{wake phase}
    \State sample tasks
    $\{x_1, \dots, x_k\} \sim \mathcal{T}$

    \ForAll{$x_i \in \{x_1, \dots, x_k\}$}
        \State
        $\tau_i \sim
        \actorvar{\pi}_{\libraryvar{L}}(x_i)$
        \Comment{rollout in $E$ starting from task $x_i$}
    \EndFor

    \State $B \gets \{\tau_1, \dots, \tau_k\}$

    \State
    $\historyvar{H}
    \gets
    \historyvar{H}
    \cup
    \{(B,\libraryvar{L})\}$
    \Comment{tag trajectories with the generating library}

    \State \Comment{sleep phase}

    \State
    $\libraryvar{L}
    \gets
    \inducervar{g}
    (\libraryvar{L},\historyvar{H})$
     \Comment{inducer inspects trajectories and past libraries in $H$; diagnoses failures/regressions; edits skills; sets public/private visibility}
\EndFor

\State \Return $\libraryvar{L}$
\end{algorithmic}
\end{algorithm}

\textbf{Intuition.}
The solution is to make the inducer a coding agent.
A coding agent has access to a code execution environment and can add, edit, and delete library skills.
If trajectories are persisted to memory or disk, the inducer can inspect failures, aggregate statistics across episodes, and test hypotheses about recurring patterns.
This lets it analyze traces surgically without placing whole trajectories in context.
Because the raw trajectories are still saved, this avoids the signal loss introduced by summary-based compression.

\textbf{Harness.}
The inducer acts as a programmer examining history and deciding what logic to implement.
It must understand the environment, diagnose failures, assign credit to skills, and check for regressions.
To support these analyses, we store both trajectories and past library versions.
Each trajectory is indexed by the
$\libraryvar{L}$ version that generated it
(line~10 of Algorithm~\ref{alg:speedrunner}), allowing the inducer
$\inducervar{g}$ to analyze how functions evolve or regress over time.
We also augment trajectories with call stacks for invoked skills: when a skill expands into a sequence of primitive actions, we record those intermediate actions and their outputs.
These call stacks help diagnose silent skill failures.

Finally, we control library size.
If the library grows without bound, the actor considers too many functions in context and performance degrades \citep{wang2025hell}.
We therefore introduce public and private access modifiers.
The inducer is instructed to make helper functions private and main functions public; private functions remain callable by other skills but are hidden from the actor.

\section{Experimental Setup}

\subsection{Baselines}
\label{baseline-selection}
We compare four methods that differ in how they accumulate experience---none, prose, or code---while standardizing the primitive actions available to both the actor and inducer across all baselines.

\textbf{ReAct} 
This baseline serves as a control with no learning between episodes. 
The actor uses only the benchmark's atomic action interface and whatever context is available within the current episode \citep{yao2023react}; it does not update any persistent strategy notes or reusable skills from past experience. 

\textbf{Online Prompt Optimization (OPO).}
We implement an online version of prompt optimization methods such as \citet{agrawal2026gepareflectivepromptevolution,yuksekgonul2025optimizing}.
After each training batch, a sleep-phase LLM reviews a concatenation of recent trajectories and updates the actor prompt.

\textbf{ASI.}
Where supported, we also compare against Agent Skill Induction (ASI) \citep{wang2025inducingprogrammaticskillsagentic}, a recent code-based baseline that induces skills from successful trajectories. Unlike \method, ASI is append-only and verifies newly induced skills through replay-style checking on the inducing task instance. To satisfy our no-replay constraint, we remove this verification step. Appendix \ref{app:asi_adapt} details this adaptation, and Appendix~\ref{app:replay_ablation} reports an ablation of its effect.

\textbf{Voyager.}
For Crafter, we additionally compare against Voyager~\citep{wang_voyager_2023}, a code-based lifelong-learning agent that learns through an automatic curriculum and an append-only executable skill library. We evaluate Voyager only on Crafter because our online protocol removes the replay-and-retry behavior available in Voyager’s original setting: once a rollout ends, the agent cannot reset to the same state, preserve inventory or other environment progress, and retry the same attempted skill. Under this constraint, Voyager’s curriculum is well defined only when the agent can continue pursuing a relatively stable objective across rollouts.

Crafter satisfies this requirement. Although each rollout starts in a new world, the high-level objective remains fixed: make progress in the survival environment. Voyager can therefore continue proposing and refining intermediate exploration goals across rollouts without the target distribution changing after every episode. ScienceWorld and BabyAI do not have this property. Their goals are externally specified and vary from episode to episode, so Voyager’s curriculum would be optimizing against a moving target. Evaluating Voyager on these benchmarks would require additional modification on the benchmark itself, making the comparison less faithful to the original algorithm. We therefore report Voyager only on Crafter, where a small adaptation is sufficient to fit our online setting; details are provided in Appendix~\ref{app:voyager_adapt}.

\subsection{Selected Benchmarks}

We evaluate on three text-based embodied benchmarks---ScienceWorld, BabyAI, and Crafter. Each defines its own physics, action space, and procedural demands, so success across all three is stronger evidence of generality than success across benchmarks that share a single underlying medium (e.g., compute-use benchmarks where the primitives reduce to file and terminal operations regardless of the task). We select subtasks where ReAct is neither trivial nor saturated, allowing us to isolate gains from skill induction.

\subsubsection{ScienceWorld}

ScienceWorld \citep{wang-etal-2022-scienceworld} is a text-based interactive environment grounded in elementary school science curricula with 10 task categories and 25 typed atomic actions (full list in Appendix~\ref{app:scienceworld}).
An episode succeeds when the simulator score reaches $1.0$; partial scores are binarized to failure.

Since skill reuse happens within category, we evaluate one category at a time. We choose \texttt{Electricity} (task~3) and \texttt{Classification} (task~4) because they are (a) solvable but not saturated by the out-of-the-box agent and (b) have enough tasks to measure learning over time.

\subsubsection{BabyAI}

BabyAI \citep{chevalier-boisvert2018babyai} is a procedurally generated 2D grid-world for instruction following under partial observability. We use the BabyAI-Text wrapper \citep{cartagrounding} as integrated in BALROG \citep{paglieri2025balrog}, which surfaces text observations. The agent acts through 6 primitives; episodes are scored as binary success.

Of BALROG's 5 subtasks, we evaluate on \texttt{pick\_up\_seq\_go\_to}, which requires sequentially picking up an object and navigating to a goal in the correct order. The ReAct baseline already saturates the other four, leaving no headroom to measure gains from skill induction. We learn from 200 training episodes (full details in Appendix~\ref{app:babyai}).

\subsubsection{Crafter}

Crafter \citep{crafter} is an open-ended survival benchmark inspired by Minecraft, where agents gather resources, craft tools, and survive hazards in procedurally generated worlds.
We expose 18 primitive actions and score each episode by the fraction of 22 predefined achievements unlocked before death or the 2{,}000-action budget.
Unlike the BALROG \citep{paglieri2025balrog} text wrapper, we expose full episode history and a richer textual view of the surroundings (full details in Appendix~\ref{app:crafter}).

\subsection{Evaluation Setup}

All experiments use \texttt{gpt-5.4-mini} for both the actor and inducer. Each run consists of $200$ online rollouts with a sleep cycle every $10$ rollouts. 
We impose a 30-minute wall-clock limit on each rollout to prevent nonterminating or pathologically long executions, such as loops inside generated skills. When the limit is reached, the rollout is terminated and scored using the environment state at termination; its token usage and partial task progress are retained in the reported metrics.
All settings are repeated across $3$ random seeds; we report means and $\pm 1$ standard deviation across seed-level results to reflect run-to-run variability. A fixed held-out test set of $30$ episodes per benchmark is shared across all methods and checkpoints; evaluation is performed every $50$ training rollouts. We evaluate along two axes: task progression (benchmark-specific success or achievement rate) and efficiency (output tokens per episode). Full implementation details are provided in Appendix~\ref{app:implementation_details_all_envs}.

\section{Results}\label{sec:main-results}

\begin{figure*}[t!]
    \centering
    \includegraphics[width=\textwidth]{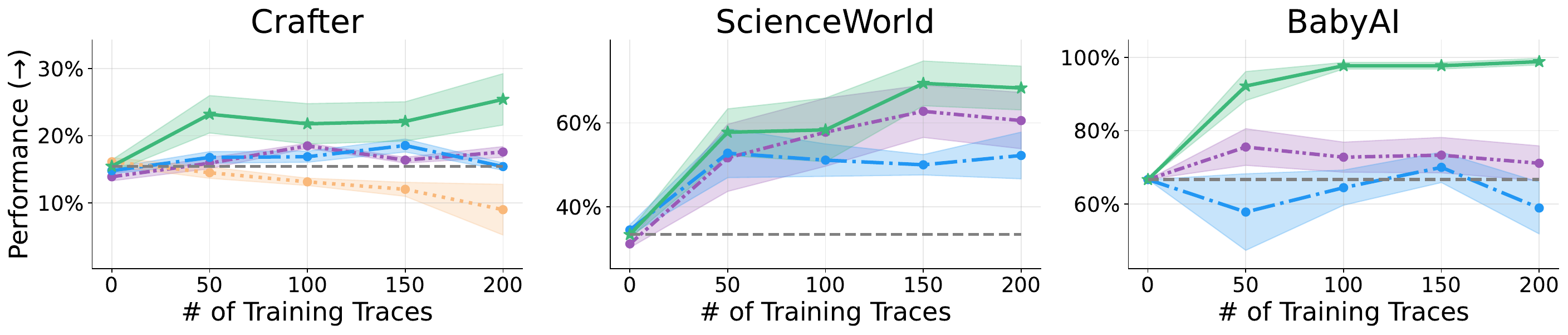}

    \vspace{0.4em}

    \includegraphics[width=\textwidth]{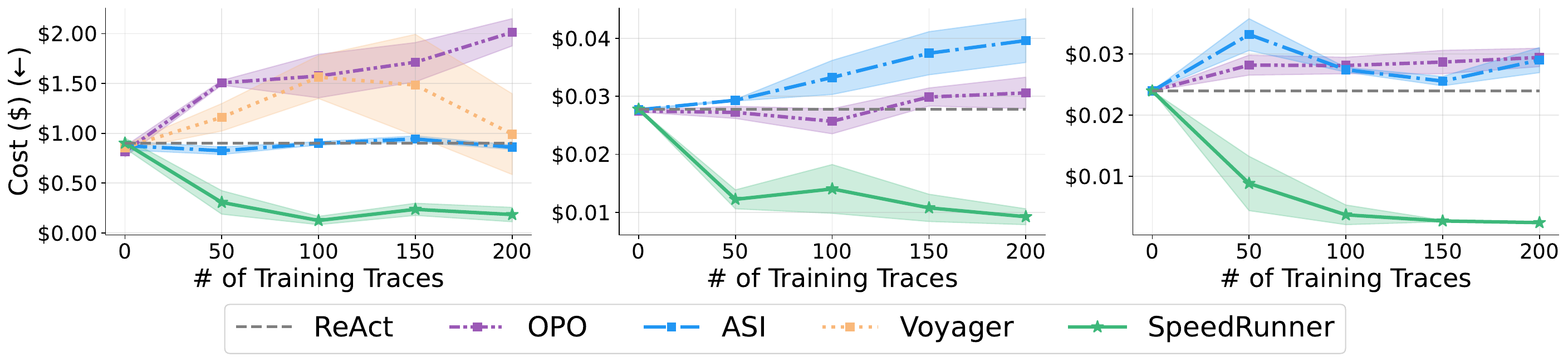}

    \caption{
    Performance and output-token cost across Crafter, ScienceWorld, and BabyAI.
    Our method and all baselines use GPT-5.4-mini as their backbone LLM.
    When the library overflows the model context limit, as happened for ASI, OPO and Voyager in Crafter, the sleep phase produces no changes and the process continues.
    \textbf{\method significantly outperforms all baselines (\S\ref{baseline-selection}) across benchmarks in terms of performance and cost except for OPO in ScienceWorld on performance, per two-sided paired t-tests.}
    }
    \label{fig:combined_performance}
\end{figure*}

Across all three benchmarks, \method improves the performance--cost tradeoff of online skill induction. The clearest and most consistent effect is efficiency: \method substantially reduces cost on every benchmark. Performance also improves, but less uniformly, suggesting that the value of executable skill induction depends on how much reusable behavior the environment exposes. All results below use GPT-5.4-mini, however, our cross-model experiments using Gemini-3-Flash and Qwen-3.5-27B in Appendix~\ref{app:gemini_results} reveal that our method's performance--cost tradeoff improvements hold in most settings.

\textbf{Performance.}
\method\ achieves the strongest final performance on all three benchmarks, with the largest margins in domains where learned routines transfer reliably across episodes. On BabyAI, \method\ climbs from the $\sim$$67\%$ ReAct baseline to near-perfect performance, a regime no other method reaches. On ScienceWorld, \method\ ties OPO at the top and substantially outperforms ASI.  Crafter is the harder case (every method ends below $30\%$ mean progression) but \method\ still leads. Appendix~\ref{app:per-benchmark-gains} discusses why the magnitude of these gains varies with benchmark structure.

\textbf{Efficiency.}
\method\ is the only method whose cost \emph{decreases}
over training, and the reduction holds across all three benchmarks --- most dramatically on BabyAI, where final usage falls to roughly an eighth of the ReAct baseline. This drop indicates that the library compresses repeated behavior into callable routines: the actor invokes high-level skills rather than re-deriving primitive sequences. ASI moves in the opposite direction on every benchmark, with cost growing as the library accumulates.  OPO stays close to the ReAct baseline on ScienceWorld and BabyAI but grows substantially on Crafter, suggesting that prose memory cannot compress recurring patterns into reusable structure when the environment is high-entropy. 

\begin{figure*}[t]
    \centering
    \includegraphics[width=\textwidth]{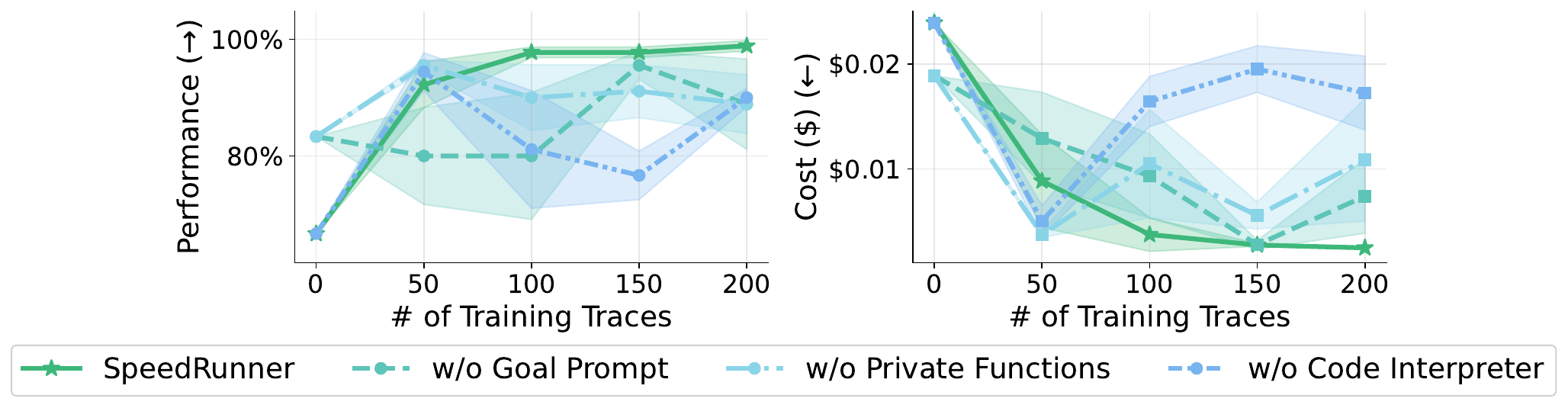}
    \caption{Ablating \method components on the BabyAI environment. Individual components have only small effects on \method's performance, however, removing the code interpreter degrades efficiency considerably.}
    \label{fig:babyai_ablations}
\end{figure*}

\section{Discussion}

\subsection{Ablation Study}

We ablate three components of \method on BabyAI: the goal-oriented inducer prompt, the public/private function split, and the inducer's code interpreter. 
The first ablation uses an ASI-style prompt that abstracts observed behavior rather than directly optimizing task solution; the second exposes all helper functions to the actor; the third replaces code-assisted trace analysis with in-context trajectory inspection. 
Due to experiment cost, full ablations are limited to BabyAI. Further discussion on the value of the code interpreter across environments can be found in Appendix \ref{app:code_interpreter_ablation}. 

Figure~\ref{fig:babyai_ablations} shows that all three ablations reduce performance, increase token usage, and introduce instability.
Removing the code interpreter has the largest efficiency cost, suggesting that programmatic trace analysis is especially important for compression.

\subsection{Code Complexity Analysis}

\providecommand{\sd}[1]{\scriptsize$\pm$#1}
\providecommand{\footstar}{\textsuperscript{$\star$}}

\subsubsection{Quantitative Analysis}
We summarize the main quantitative findings here and defer the full structural statistics to Appendix~\ref{app:quantitative}. For each final codebook, we construct a directed call graph $G=(V,E)$, where nodes are induced functions and edges denote function calls. We focus on two call-graph measures in Table~\ref{tab:complexity_summary}: maximum depth and density. Maximum depth is the length of the longest directed path in $G$, measuring the deepest abstraction chain in the library. Density is $|E|/(|V|(|V|-1))$, excluding self-edges, and measures how often induced functions call one another rather than only primitive actions.

\paragraph{Deeper and denser call graphs.}
Table~\ref{tab:complexity_summary} shows that \method\ learns more structurally organized codebooks than the baselines. Across benchmark settings, \method\ generally produces deeper and denser call graphs than ASI. This indicates that its learned skills are not merely independent wrappers around primitive actions, but are arranged into reusable hierarchical routines.

\paragraph{Reuse rather than accumulation.}
The contrast is especially clear on Crafter, where all methods are available. Voyager reaches maximum depth $5.7$, but its density is only $0.0005$, indicating little reuse among induced functions. In comparison, \method\ reaches greater maximum depth ($8.7$) and much higher density ($0.149$). Thus, \method\ does not improve by accumulating many scenario-specific functions; it factors repeated behavior into shared abstractions that later skills can call.

Overall, the quantitative signature of \method\ is compositional organization: its functions call one another more often and form deeper abstraction hierarchies.

\begin{table*}[t]
\centering
\small
\setlength{\tabcolsep}{5pt}
\begin{tabular}{lcc cc cc}
\toprule
\multirow{2}{*}{\textbf{Method}}
& \multicolumn{2}{c}{\textbf{Crafter}}
& \multicolumn{2}{c}{\textbf{ScienceWorld}}
& \multicolumn{2}{c}{\textbf{BabyAI}} \\
\cmidrule(lr){2-3}
\cmidrule(lr){4-5}
\cmidrule(lr){6-7}
& \textbf{Max d.} & \textbf{Density}
& \textbf{Max d.} & \textbf{Density}
& \textbf{Max d.} & \textbf{Density} \\
\midrule
ASI
& $1.7${\sd{0.6}} & $0.030${\sd{0.008}}
& $1.7${\sd{0.3}} & $0.016${\sd{0.004}}
& $1.0${\sd{0.0}} & $0.018${\sd{0.005}} \\

\method\ w/o CI
& $6.3${\sd{3.2}} & $0.138${\sd{0.078}}
& $5.2${\sd{1.2}} & $\bm{0.073}${\sd{0.013}}
& $\bm{3.3}${\sd{1.5}} & $0.178${\sd{0.019}} \\

\rowcolor{oursrow}
\textbf{\method\ (ours)}
& $\bm{8.7}${\sd{2.1}} & $\bm{0.149}${\sd{0.042}}
& $\bm{5.8}${\sd{0.8}} & $0.067${\sd{0.018}}
& $2.3${\sd{0.6}} & $\bm{0.206}${\sd{0.042}} \\

Voyager
& $5.7${\sd{2.1}} & $0.0005${\sd{0.0003}}
& -- & --
& -- & -- \\
\bottomrule
\end{tabular}
\caption{Main call-graph statistics for final skill libraries, using GPT-5.4-mini with mean$\pm$sd over three seeds. The ScienceWorld column averages task~3 and task~4; reported means and standard deviations are averaged across the two tasks. We report maximum call-graph depth (Max d.) and call-graph density. \textbf{\method learns deeper and denser call graphs.}}
\label{tab:complexity_summary}
\end{table*}

\subsubsection{Qualitative Analysis}

To understand where \method's gains come from, we now qualitatively compare our best-trial libraries against those of our baselines and ablations. Two patterns recur across all four environments; we summarize them here and defer per-environment evidence to Appendix~\ref{app:qualitative}.

\textbf{Code-enabled trace filtering.} The inducer's code interpreter lets it query specific slices of actor trajectories before committing a function. In BabyAI, candidate mission parsers were executed against historical mission strings to reveal that reversed-order phrasings (\emph{``go to X after you pick up Y''}) invert execution order---a failure mode that the ablation's LLM-only inducer misses on 6.7\% of episodes. In Crafter, the inducer noticed a high rate of \texttt{not\_found} returns from \texttt{travel\_to\_visible\_resource}, queried trace history to confirm the function was being called hundreds of times per batch while still failing, and introduced \texttt{\_bounded\_exploration} to sweep the local neighborhood for the desired resource before giving up. Without this loop, fixes accrue reactively one failure at a time.

\textbf{Higher-level skills.} The code interpreter enables abstractions that span high-level goals rather than a single primitive, a category of function essentially unique to \method. ASI and Voyager produce primitives templated from individual traces (fixed motion sequences, per-resource collectors), with no high level goal logic. In contrast, \method generates functions such as \texttt{solve\_sequential\_mission} (BabyAI), \texttt{\_find\_visible\_name\_candidates} (ScienceWorld), and \texttt{\_choose\_progress\_target} (Crafter), each encoding a complex decision hierarchy in one place. The actor can often complete episodes with one or two high-level calls, explaining \method's 2--8$\times$ token reduction over ASI.
 
\subsection{Effect of External Randomness (Crafter)}
\label{sec:analysis:external-randomness}

A central concern with code-based skill induction is that induced
functions are \emph{deterministic} by construction and therefore
potentially brittle to environmental noise: each function commits to a
fixed sequence of primitive actions, while real environments rarely
repeat themselves exactly.  We refer to this as the \emph{external
randomness gap}.  Crafter lets us isolate this factor by varying the
zombie spawn frequency --- \textbf{0x} (no zombies), \textbf{1x}
(standard), \textbf{2x} (doubled) --- while holding all other mechanics
fixed.

\begin{figure}[t]
\centering
\includegraphics[width=0.6\columnwidth]{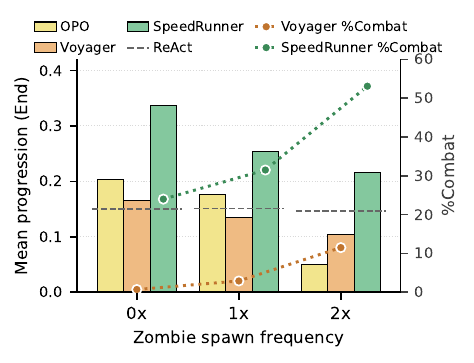}
\caption{Crafter final performance vs \emph{\%Combat}, the fraction of induced functions whose body references a creature/combat keyword under different zombie settings (GPT-5.4-mini, 200 training
rollouts).}
\label{fig:zombie_sweep}
\end{figure}

Figure~\ref{fig:zombie_sweep} (bars) evaluates the robustness of all
three methods to external randomness, measured against the ReAct
baseline.  Prose memory (OPO) remains useful under low randomness, but collapses once the environment becomes sufficiently stochastic.
Append-only code (Voyager) shows a similar trend.  In contrast, only
\method\ improves over the baseline in every condition.

The \%Combat overlay (Figure~\ref{fig:zombie_sweep}, dotted lines)
characterizes the learned codebooks.  As zombie pressure increases,
\method\ does not simply expand its library; it reallocates
representational budget based on the noisy signal, becoming deeper and
more combat-aware by revising existing skills in place.  This allows a
single defensive abstraction to be amortized across the call graph.
Voyager, by contrast, repeatedly reintroduces the same combat logic as
duplicated branches inside newly generated, encounter-specific
functions, without consolidating it into a general abstraction
(Appendix~\ref{app:zombie_library_table} gives the detailed per-condition
structural breakdown and examples).

Together, these results suggest that \method\ is better able to
extract useful learning signal from stochastic traces.  Rather than
treating external randomness as irreducible noise, it identifies
recurring failure modes and quickly incorporates the corresponding
adaptations into reusable abstractions.

\subsection{Adaptation under Distribution Shift (ScienceWorld)} \label{sec:distribution-shift}
\begin{figure*}[t]
    \centering
    \includegraphics[width=0.9\textwidth]{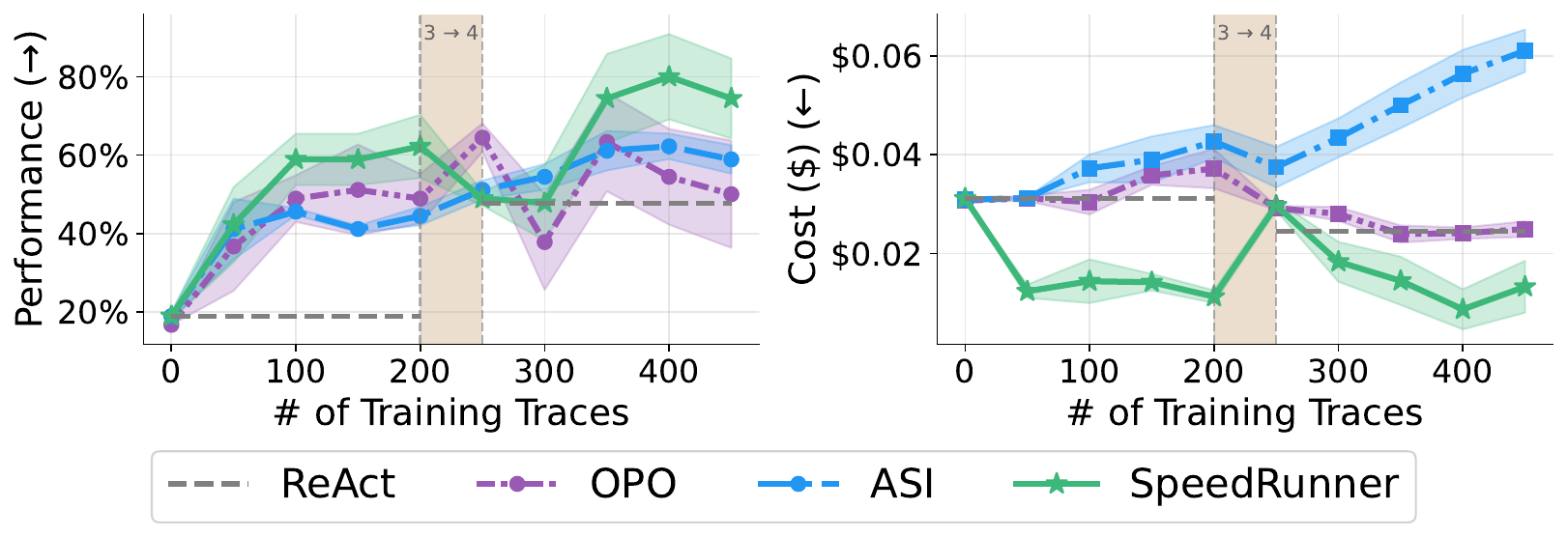}
    \caption{ScienceWorld distribution shift: performance (left) and cost (right) when switching from task~3 to task~4 midway through training. \textbf{\method achieves the best performance and strongest token compression across both tasks.}
        }
    \label{fig:scienceworld_shift}
\end{figure*}

Agents deployed in long-running environments must continually adapt to changing task distributions. A central challenge is to acquire new capabilities while retaining previously acquired ones. We study this setting by training on ScienceWorld task~3 (\texttt{Electricity}) for 200 rollouts before switching to task~4 (\texttt{Classification}), carrying the library over unchanged (Figure~\ref{fig:scienceworld_shift}; GPT-5.4-mini, three seeds).

At the distribution shift, all methods’ performance and cost approach ReAct’s behavior because the two tasks require largely disjoint procedural knowledge. Throughout learning, \method achieves the best performance and strongest token compression on both tasks. By the end of task~4. it refines a compact library of approximately 40 functions, compared to ASI which accumulates approximately 160 functions and plateaus at lower performance. 

To evaluate knowledge retention, we re-evaluate task~3 after task~4 training. Despite actively revising its library, \method loses only 3.3 percentage points on task~3 while reducing inference cost by 27.1\%, making it the only method that becomes more efficient on the original task after adapting to a new distribution. In contrast, ASI improves task~3 performance by 4.4 percentage points but increases inference cost by 40.6\% because of its append-only nature: it preserves all learned knowledge, at the cost of unbounded library growth. OPO loses 5.6 percentage points on task~3 while increasing inference cost by 6.8\%, performing worse than \method in both retention and efficiency.

\section{Related Work}

Our work builds on the broader literature on continual learning \citep{Kirkpatrick_2017, li2017learningforgetting, wang2024comprehensivesurveycontinuallearning, shi2024continuallearninglargelanguage} and self-evolving agents \citep{fang2025comprehensivesurveyselfevolvingai, gao2026surveyselfevolvingagentswhat}, which aim to build agents that improve continuously through interaction with their environment. Existing approaches broadly follow two lines: adapting the model's weights \citep{qu2025rladtrainingllmsdiscover, song2024trialerrorexplorationbasedtrajectory, wang2025ragenunderstandingselfevolutionllm} or augmenting a frozen model with externally stored knowledge \citep{shinn2023reflexionlanguageagentsverbal, wang_voyager_2023, zhao2024expelllmagentsexperiential}.  We study the latter, specifically \emph{skill learning}, which distills experience into reusable skills loaded into context to help complete a task. Prior skill-learning methods primarily aim to improve task performance and adaptation, while treating inference cost as a secondary outcome. We instead study skill learning through the lens of inference efficiency, where the central design choice is how skills are represented.

Most methods represent skills in natural language. A line of work distills trajectories into reusable natural-language skill documents temporarily loaded into context when relevant \citep{zhao2024expelllmagentsexperiential, wang_agent_2025, ma_skillclaw_2026, wang2026skillxautomaticallyconstructingskill, liu2026skillforgeforgingdomainspecificselfevolving, ni2026trace2skilldistilltrajectorylocallessons, alzubi2026evoskillautomatedskilldiscovery}. Some methods adopt skill folders which bundle textual instructions and helper scripts \citep{liu2026skillforgeforgingdomainspecificselfevolving, alzubi2026evoskillautomatedskilldiscovery}, but the primary optimization target remains the natural-language artifact. While natural-language skills are flexible and generalize well, the agent must reread and re-follow the instructions on every reuse, so recurring procedures keep incurring inference cost.

The second line of work represents reusable skills as executable code. Foundational works used Bayesian program induction to synthesize formal representations through bottom-up composition of primitives \citep{ellis_library_2018, ellis_dreamcoder_2021}. Recent work extends this paradigm to LLM agents: Voyager grows a skill library through open-ended Minecraft exploration \citep{wang_voyager_2023}, while a wave of systems induce and refine skills from agent trajectories \citep{stengel-eskin_regal_2024, wang2025inducingprogrammaticskillsagentic, zheng2025skillweaverwebagentsselfimprove, qiu2026autorefinetrajectoriesreusableexpertise}. Programmatic skills replace repeated reasoning with delegated computation and thus provide a more efficient representation for recurring procedures. However, previous work primarily studies programmatic skills to improve agent capability. Many methods further rely on experience replay for validating candidate skills to guard against regression, yet these assumptions are difficult to satisfy in deployed environments where interactions are irreversible.

Our work similarly represents skills as executable code but differs in both objective and methodology. Rather than treating programmatic skills primarily as a means of improving capability, we highlight their capacity to minimize inference cost by maximizing reusable computation. We further show that such skills can be induced entirely online without replay by keeping the accumulated trajectories available as a growing dataset for programmatic analysis, enabling continual refinement of the skill library while mitigating regression.

\section{Conclusion}

In this work, we demonstrate that representing skills as code can achieve the best cost effectiveness while retaining performance. 
We observe that this trend is remarkably consistent across distribution shifts, randomness, different domains, and even different models. 
We further demonstrate that this trend exists in a purely-online setting, without a held-out set of episodes to evaluate skill update quality for hillclimbing. 
Not all programmatic skill learning approaches reduce cost to the same extent.
We are only able to do so by framing the process of extracting signal from past trajectories as an agentic coding task.

\textbf{Limitations}
We were unable to benchmark flagship models due to cost limitations.
For the same reason, we were unable to evaluate on more expensive benchmarks such as SWE-Bench \citep{jimenez2024swebench} or Terminal-Bench \citep{merrill2026terminalbench}.
Nevertheless, we believe that the diversity and controllability of our evaluations are sufficient to support our general conclusions.
Skill learning at test time suffers from instability, as demonstrated by the large error bars across all methods we evaluated; stabilizing online skill induction remains an open problem.
Lastly, we do not study meta-learning of the inducer itself---in principle, the inducer could learn to perform better analyses over time, but we leave this to future work.

\bibliography{biblio}

@inproceedings{wang_agent_2025,
    title = {Agent {Workflow} {Memory}},
    url = {https://openreview.net/forum?id=NTAhi2JEEE},
    language = {en},
    urldate = {2026-05-03},
    author = {Wang, Zora Zhiruo and Mao, Jiayuan and Fried, Daniel and Neubig, Graham},
    month = jun,
    year = {2025},
}

@misc{ma_skillclaw_2026,
    title = {{SkillClaw}: {Let} {Skills} {Evolve} {Collectively} with {Agentic} {Evolver}},
    shorttitle = {{SkillClaw}},
    url = {http://arxiv.org/abs/2604.08377},
    doi = {10.48550/arXiv.2604.08377},
    urldate = {2026-05-03},
    publisher = {arXiv},
    author = {Ma, Ziyu and Yang, Shidong and Ji, Yuxiang and Wang, Xucong and Wang, Yong and Hu, Yiming and Huang, Tongwen and Chu, Xiangxiang},
    month = apr,
    year = {2026},
    note = {arXiv:2604.08377 [cs]},
}

@article{wang_voyager_2023,
    title = {Voyager: {An} {Open}-{Ended} {Embodied} {Agent} with {Large} {Language} {Models}},
    issn = {2835-8856},
    shorttitle = {Voyager},
    url = {https://openreview.net/forum?id=ehfRiF0R3a},
    language = {en},
    urldate = {2026-05-03},
    journal = {Transactions on Machine Learning Research},
    author = {Wang, Guanzhi and Xie, Yuqi and Jiang, Yunfan and Mandlekar, Ajay and Xiao, Chaowei and Zhu, Yuke and Fan, Linxi and Anandkumar, Anima},
    month = nov,
    year = {2023},
}

@inproceedings{ellis_library_2018,
    address = {Red Hook, NY, USA},
    series = {{NIPS}'18},
    title = {Library learning for neurally-guided {Bayesian} program induction},
    url = {https://dl.acm.org/doi/10.5555/3327757.3327878},
    urldate = {2026-05-03},
    booktitle = {Proceedings of the 32nd {International} {Conference} on {Neural} {Information} {Processing} {Systems}},
    publisher = {Curran Associates Inc.},
    author = {Ellis, Kevin and Morales, Lucas and Sablé-Meyer, Mathias and Solar-Lezama, Armando and Tenenbaum, Joshua B.},
    month = dec,
    year = {2018},
    pages = {7816--7826},
}

@inproceedings{ellis_dreamcoder_2021,
    address = {New York, NY, USA},
    series = {{PLDI} 2021},
    title = {{DreamCoder}: bootstrapping inductive program synthesis with wake-sleep library learning},
    isbn = {978-1-4503-8391-2},
    shorttitle = {{DreamCoder}},
    url = {https://dl.acm.org/doi/10.1145/3453483.3454080},
    doi = {10.1145/3453483.3454080},
    urldate = {2026-05-03},
    booktitle = {Proceedings of the 42nd {ACM} {SIGPLAN} {International} {Conference} on {Programming} {Language} {Design} and {Implementation}},
    publisher = {Association for Computing Machinery},
    author = {Ellis, Kevin and Wong, Catherine and Nye, Maxwell and Sablé-Meyer, Mathias and Morales, Lucas and Hewitt, Luke and Cary, Luc and Solar-Lezama, Armando and Tenenbaum, Joshua B.},
    month = jun,
    year = {2021},
    pages = {835--850},
}

@misc{stengel-eskin_regal_2024,
    title = {{ReGAL}: {Refactoring} {Programs} to {Discover} {Generalizable} {Abstractions}},
    shorttitle = {{ReGAL}},
    url = {http://arxiv.org/abs/2401.16467},
    doi = {10.48550/arXiv.2401.16467},
    urldate = {2024-07-09},
    publisher = {arXiv},
    author = {Stengel-Eskin, Elias and Prasad, Archiki and Bansal, Mohit},
    month = jun,
    year = {2024},
    note = {arXiv:2401.16467 [cs]},
}

@misc{wang2025inducingprogrammaticskillsagentic,
      title={Inducing Programmatic Skills for Agentic Tasks}, 
      author={Zora Zhiruo Wang and Apurva Gandhi and Graham Neubig and Daniel Fried},
      year={2025},
      eprint={2504.06821},
      archivePrefix={arXiv},
      primaryClass={cs.CL},
      url={https://arxiv.org/abs/2504.06821}, 
}

@misc{zheng2025skillweaverwebagentsselfimprove,
      title={SkillWeaver: Web Agents can Self-Improve by Discovering and Honing Skills}, 
      author={Boyuan Zheng and Michael Y. Fatemi and Xiaolong Jin and Zora Zhiruo Wang and Apurva Gandhi and Yueqi Song and Yu Gu and Jayanth Srinivasa and Gaowen Liu and Graham Neubig and Yu Su},
      year={2025},
      eprint={2504.07079},
      archivePrefix={arXiv},
      primaryClass={cs.AI},
      url={https://arxiv.org/abs/2504.07079}, 
}

@misc{agrawal2026gepareflectivepromptevolution,
      title={GEPA: Reflective Prompt Evolution Can Outperform Reinforcement Learning}, 
      author={Lakshya A Agrawal and Shangyin Tan and Dilara Soylu and Noah Ziems and Rishi Khare and Krista Opsahl-Ong and Arnav Singhvi and Herumb Shandilya and Michael J Ryan and Meng Jiang and Christopher Potts and Koushik Sen and Alexandros G. Dimakis and Ion Stoica and Dan Klein and Matei Zaharia and Omar Khattab},
      year={2026},
      eprint={2507.19457},
      archivePrefix={arXiv},
      primaryClass={cs.CL},
      url={https://arxiv.org/abs/2507.19457}, 
}

@misc{pryzant2023automaticpromptoptimizationgradient,
      title={Automatic Prompt Optimization with "Gradient Descent" and Beam Search}, 
      author={Reid Pryzant and Dan Iter and Jerry Li and Yin Tat Lee and Chenguang Zhu and Michael Zeng},
      year={2023},
      eprint={2305.03495},
      archivePrefix={arXiv},
      primaryClass={cs.CL},
      url={https://arxiv.org/abs/2305.03495}, 
}

@inproceedings{
wang2025hell,
title={Hell or High Water: Evaluating Agentic Recovery from External Failures},
author={Andrew Wang and Sophia Hager and Adi Asija and Daniel Khashabi and Nicholas Andrews},
booktitle={Second Conference on Language Modeling},
year={2025},
url={https://openreview.net/forum?id=Zk224WPT42}
}

@inproceedings{wang-etal-2022-scienceworld,
    title = "{S}cience{W}orld: Is your Agent Smarter than a 5th Grader?",
    author = "Wang, Ruoyao  and
      Jansen, Peter  and
      C{\^o}t{\'e}, Marc-Alexandre  and
      Ammanabrolu, Prithviraj",
    editor = "Goldberg, Yoav  and
      Kozareva, Zornitsa  and
      Zhang, Yue",
    booktitle = "Proceedings of the 2022 Conference on Empirical Methods in Natural Language Processing",
    month = dec,
    year = "2022",
    address = "Abu Dhabi, United Arab Emirates",
    publisher = "Association for Computational Linguistics",
    url = "https://aclanthology.org/2022.emnlp-main.775/",
    doi = "10.18653/v1/2022.emnlp-main.775",
    pages = "11279--11298"
}

@inproceedings{
crafter,
title={Benchmarking the Spectrum of Agent Capabilities},
author={Danijar Hafner},
booktitle={International Conference on Learning Representations},
year={2022},
url={https://openreview.net/forum?id=1W0z96MFEoH}
}

@inproceedings{
chevalier-boisvert2018babyai,
title={Baby{AI}: First Steps Towards Grounded Language Learning With a Human In the Loop},
author={Maxime Chevalier-Boisvert and Dzmitry Bahdanau and Salem Lahlou and Lucas Willems and Chitwan Saharia and Thien Huu Nguyen and Yoshua Bengio},
booktitle={International Conference on Learning Representations},
year={2019},
url={https://openreview.net/forum?id=rJeXCo0cYX},
}

@inproceedings{
yao2023react,
title={ReAct: Synergizing Reasoning and Acting in Language Models},
author={Shunyu Yao and Jeffrey Zhao and Dian Yu and Nan Du and Izhak Shafran and Karthik R Narasimhan and Yuan Cao},
booktitle={The Eleventh International Conference on Learning Representations },
year={2023},
url={https://openreview.net/forum?id=WE_vluYUL-X}
}

@inproceedings{
jimenez2024swebench,
title={{SWE}-bench: Can Language Models Resolve Real-world Github Issues?},
author={Carlos E Jimenez and John Yang and Alexander Wettig and Shunyu Yao and Kexin Pei and Ofir Press and Karthik R Narasimhan},
booktitle={The Twelfth International Conference on Learning Representations},
year={2024},
url={https://openreview.net/forum?id=VTF8yNQM66}
}

@inproceedings{
merrill2026terminalbench,
title={Terminal-Bench: Benchmarking Agents on Hard, Realistic Tasks in Command Line Interfaces},
author={Mike A Merrill and Alexander Glenn Shaw and Nicholas Carlini and Boxuan Li and Harsh Raj and Ivan Bercovich and Lin Shi and Jeong Yeon Shin and Thomas Walshe and E. Kelly Buchanan and Junhong Shen and Guanghao Ye and Haowei Lin and Jason Poulos and Maoyu Wang and Marianna Nezhurina and Di Lu and Orfeas Menis Mastromichalakis and Zhiwei Xu and Zizhao Chen and Yue Liu and Robert Zhang and Leon Liangyu Chen and Anurag Kashyap and Jan-Lucas Uslu and Jeffrey Li and Jianbo Wu and Minghao Yan and Song Bian and Vedang Sharma and Ke Sun and Steven Dillmann and Akshay Anand and Andrew Lanpouthakoun and Bardia Koopah and Changran Hu and Etash Kumar Guha and Gabriel H. S. Dreiman and Jiacheng Zhu and Karl Krauth and Li Zhong and Niklas Muennighoff and Robert Kwesi Amanfu and Shangyin Tan and Shreyas Pimpalgaonkar and Tushar Aggarwal and Xiangning Lin and Xin Lan and Xuandong Zhao and Yiqing Liang and Yuanli Wang and Zilong Wang and Changzhi Zhou and David Heineman and Hange Liu and Harsh Trivedi and John Yang and Junhong Lin and Manish Shetty and Michael Yang and Nabil Omi and Negin Raoof and Shanda Li and Terry Yue Zhuo and Wuwei Lin and Yiwei Dai and Yuxin Wang and Wenhao Chai and Shang Zhou and Dariush Wahdany and Ziyu She and Jiaming Hu and Zhikang Dong and Yuxuan Zhu and Sasha Cui and Ahson Saiyed and Arinbj{\"o}rn Kolbeinsson and Christopher Michael Rytting and Ryan Marten and Yixin Wang and Jenia Jitsev and Alex Dimakis and Andy Konwinski and Ludwig Schmidt},
booktitle={The Fourteenth International Conference on Learning Representations},
year={2026},
url={https://openreview.net/forum?id=a7Qa4CcHak}
}

@article{yuksekgonul2025optimizing,
  title={Optimizing generative AI by backpropagating language model feedback},
  author={Yuksekgonul, Mert and Bianchi, Federico and Boen, Joseph and Liu, Sheng and Lu, Pan and Huang, Zhi and Guestrin, Carlos and Zou, James},
  journal={Nature},
  volume={639},
  pages={609--616},
  year={2025},
}

@inproceedings{cartagrounding,
author = {Carta, Thomas and Romac, Cl\'{e}ment and Wolf, Thomas and Lamprier, Sylvain and Sigaud, Olivier and Oudeyer, Pierre-Yves},
title = {Grounding large language models in interactive environments with online reinforcement learning},
year = {2023},
publisher = {JMLR.org},
booktitle = {Proceedings of the 40th International Conference on Machine Learning},
articleno = {150},
numpages = {38},
location = {Honolulu, Hawaii, USA},
series = {ICML'23}
}

@inproceedings{
paglieri2025balrog,
title={{BALROG}: Benchmarking Agentic {LLM} and {VLM} Reasoning On Games},
author={Davide Paglieri and Bart{\l}omiej Cupia{\l} and Samuel Coward and Ulyana Piterbarg and Maciej Wolczyk and Akbir Khan and Eduardo Pignatelli and {\L}ukasz Kuci{\'n}ski and Lerrel Pinto and Rob Fergus and Jakob Nicolaus Foerster and Jack Parker-Holder and Tim Rockt{\"a}schel},
booktitle={The Thirteenth International Conference on Learning Representations},
year={2025},
url={https://openreview.net/forum?id=fp6t3F669F}
}

@misc{ni2026trace2skilldistilltrajectorylocallessons,
      title={Trace2Skill: Distill Trajectory-Local Lessons into Transferable Agent Skills}, 
      author={Jingwei Ni and Yihao Liu and Xinpeng Liu and Yutao Sun and Mengyu Zhou and Pengyu Cheng and Dexin Wang and Erchao Zhao and Xiaoxi Jiang and Guanjun Jiang},
      year={2026},
      eprint={2603.25158},
      archivePrefix={arXiv},
      primaryClass={cs.AI},
      url={https://arxiv.org/abs/2603.25158}, 
}

@misc{qiu2026autorefinetrajectoriesreusableexpertise,
      title={AutoRefine: From Trajectories to Reusable Expertise for Continual LLM Agent Refinement}, 
      author={Libin Qiu and Zhirong Gao and Junfu Chen and Yuhang Ye and Weizhi Huang and Xiaobo Xue and Wenkai Qiu and Shuo Tang},
      year={2026},
      eprint={2601.22758},
      archivePrefix={arXiv},
      primaryClass={cs.AI},
      url={https://arxiv.org/abs/2601.22758}, 
}

@misc{wang2026skillxautomaticallyconstructingskill,
      title={SkillX: Automatically Constructing Skill Knowledge Bases for Agents}, 
      author={Chenxi Wang and Zhuoyun Yu and Xin Xie and Wuguannan Yao and Runnan Fang and Shuofei Qiao and Kexin Cao and Guozhou Zheng and Xiang Qi and Peng Zhang and Shumin Deng},
      year={2026},
      eprint={2604.04804},
      archivePrefix={arXiv},
      primaryClass={cs.CL},
      url={https://arxiv.org/abs/2604.04804}, 
}

@misc{alzubi2026evoskillautomatedskilldiscovery,
      title={EvoSkill: Automated Skill Discovery for Multi-Agent Systems}, 
      author={Salaheddin Alzubi and Noah Provenzano and Jaydon Bingham and Weiyuan Chen and Tu Vu},
      year={2026},
      eprint={2603.02766},
      archivePrefix={arXiv},
      primaryClass={cs.AI},
      url={https://arxiv.org/abs/2603.02766}, 
}

@misc{liu2026skillforgeforgingdomainspecificselfevolving,
      title={SkillForge: Forging Domain-Specific, Self-Evolving Agent Skills in Cloud Technical Support}, 
      author={Xingyan Liu and Xiyue Luo and Linyu Li and Ganghong Huang and Jianfeng Liu and Honglin Qiao},
      year={2026},
      eprint={2604.08618},
      archivePrefix={arXiv},
      primaryClass={cs.IR},
      doi={https://doi.org/10.1145/3805712.3808466},
      url={https://arxiv.org/abs/2604.08618}, 
}

@misc{zhong2026skilllearnbenchbenchmarkingcontinuallearning,
      title={SkillLearnBench: Benchmarking Continual Learning Methods for Agent Skill Generation on Real-World Tasks}, 
      author={Shanshan Zhong and Yi Lu and Jingjie Ning and Yibing Wan and Lihan Feng and Yuyi Ao and Leonardo F. R. Ribeiro and Markus Dreyer and Sean Ammirati and Chenyan Xiong},
      year={2026},
      eprint={2604.20087},
      archivePrefix={arXiv},
      primaryClass={cs.CL},
      url={https://arxiv.org/abs/2604.20087}, 
}

@misc{qu2025rladtrainingllmsdiscover,
      title={RLAD: Training LLMs to Discover Abstractions for Solving Reasoning Problems}, 
      author={Yuxiao Qu and Anikait Singh and Yoonho Lee and Amrith Setlur and Ruslan Salakhutdinov and Chelsea Finn and Aviral Kumar},
      year={2025},
      eprint={2510.02263},
      archivePrefix={arXiv},
      primaryClass={cs.AI},
      url={https://arxiv.org/abs/2510.02263}, 
}

@inproceedings{
mi2026skillpro,
title={Skill-Pro: Learning Reusable Skills from Experience via Non-Parametric {PPO} for {LLM} Agents},
author={Qirui Mi and Zhijian Ma and Mengyue Yang and Haoxuan Li and Yisen Wang and Haifeng Zhang and Jun Wang},
booktitle={Forty-third International Conference on Machine Learning},
year={2026},
url={https://openreview.net/forum?id=9kJQjx2B80}
}

@misc{wang2026skillsdskillconditionedselfdistillationmultiturn,
      title={Skill-SD: Skill-Conditioned Self-Distillation for Multi-turn LLM Agents}, 
      author={Hao Wang and Guozhi Wang and Han Xiao and Yufeng Zhou and Yue Pan and Jichao Wang and Ke Xu and Yafei Wen and Xiaohu Ruan and Xiaoxin Chen and Honggang Qi},
      year={2026},
      eprint={2604.10674},
      archivePrefix={arXiv},
      primaryClass={cs.LG},
      url={https://arxiv.org/abs/2604.10674}, 
}

@misc{yang2026autoskillexperiencedrivenlifelonglearning,
      title={AutoSkill: Experience-Driven Lifelong Learning via Skill Self-Evolution}, 
      author={Yutao Yang and Junsong Li and Qianjun Pan and Bihao Zhan and Yuxuan Cai and Lin Du and Jie Zhou and Kai Chen and Qin Chen and Xin Li and Bo Zhang and Liang He},
      year={2026},
      eprint={2603.01145},
      archivePrefix={arXiv},
      primaryClass={cs.AI},
      url={https://arxiv.org/abs/2603.01145}, 
}

@article{Kirkpatrick_2017,
   title={Overcoming catastrophic forgetting in neural networks},
   volume={114},
   ISSN={1091-6490},
   url={http://dx.doi.org/10.1073/pnas.1611835114},
   DOI={10.1073/pnas.1611835114},
   number={13},
   journal={Proceedings of the National Academy of Sciences},
   publisher={National Academy of Sciences},
   author={Kirkpatrick, James and Pascanu, Razvan and Rabinowitz, Neil and Veness, Joel and Desjardins, Guillaume and Rusu, Andrei A. and Milan, Kieran and Quan, John and Ramalho, Tiago and Grabska-Barwinska, Agnieszka and Hassabis, Demis and Clopath, Claudia and Kumaran, Dharshan and Hadsell, Raia},
   year={2017},
   month=Mar, pages={3521–3526} }

@misc{li2017learningforgetting,
      title={Learning without Forgetting}, 
      author={Zhizhong Li and Derek Hoiem},
      year={2017},
      eprint={1606.09282},
      archivePrefix={arXiv},
      primaryClass={cs.CV},
      url={https://arxiv.org/abs/1606.09282}, 
}

@misc{wang2024comprehensivesurveycontinuallearning,
      title={A Comprehensive Survey of Continual Learning: Theory, Method and Application}, 
      author={Liyuan Wang and Xingxing Zhang and Hang Su and Jun Zhu},
      year={2024},
      eprint={2302.00487},
      archivePrefix={arXiv},
      primaryClass={cs.LG},
      url={https://arxiv.org/abs/2302.00487}, 
}

@misc{shi2024continuallearninglargelanguage,
      title={Continual Learning of Large Language Models: A Comprehensive Survey}, 
      author={Haizhou Shi and Zihao Xu and Hengyi Wang and Weiyi Qin and Wenyuan Wang and Yibin Wang and Zifeng Wang and Sayna Ebrahimi and Hao Wang},
      year={2024},
      eprint={2404.16789},
      archivePrefix={arXiv},
      primaryClass={cs.LG},
      url={https://arxiv.org/abs/2404.16789}, 
}

@misc{shinn2023reflexionlanguageagentsverbal,
      title={Reflexion: Language Agents with Verbal Reinforcement Learning}, 
      author={Noah Shinn and Federico Cassano and Edward Berman and Ashwin Gopinath and Karthik Narasimhan and Shunyu Yao},
      year={2023},
      eprint={2303.11366},
      archivePrefix={arXiv},
      primaryClass={cs.AI},
      url={https://arxiv.org/abs/2303.11366}, 
}

@misc{zhao2024expelllmagentsexperiential,
      title={ExpeL: LLM Agents Are Experiential Learners}, 
      author={Andrew Zhao and Daniel Huang and Quentin Xu and Matthieu Lin and Yong-Jin Liu and Gao Huang},
      year={2024},
      eprint={2308.10144},
      archivePrefix={arXiv},
      primaryClass={cs.LG},
      url={https://arxiv.org/abs/2308.10144}, 
}

@misc{gao2026surveyselfevolvingagentswhat,
      title={A Survey of Self-Evolving Agents: What, When, How, and Where to Evolve on the Path to Artificial Super Intelligence}, 
      author={Huan-ang Gao and Jiayi Geng and Wenyue Hua and Mengkang Hu and Xinzhe Juan and Hongzhang Liu and Shilong Liu and Jiahao Qiu and Xuan Qi and Yiran Wu and Hongru Wang and Han Xiao and Yuhang Zhou and Shaokun Zhang and Jiayi Zhang and Jinyu Xiang and Yixiong Fang and Qiwen Zhao and Dongrui Liu and Qihan Ren and Cheng Qian and Zhenhailong Wang and Minda Hu and Huazheng Wang and Qingyun Wu and Heng Ji and Mengdi Wang},
      year={2026},
      eprint={2507.21046},
      archivePrefix={arXiv},
      primaryClass={cs.AI},
      url={https://arxiv.org/abs/2507.21046}, 
}

@misc{fang2025comprehensivesurveyselfevolvingai,
      title={A Comprehensive Survey of Self-Evolving AI Agents: A New Paradigm Bridging Foundation Models and Lifelong Agentic Systems}, 
      author={Jinyuan Fang and Yanwen Peng and Xi Zhang and Yingxu Wang and Xinhao Yi and Guibin Zhang and Yi Xu and Bin Wu and Siwei Liu and Zihao Li and Zhaochun Ren and Nikos Aletras and Xi Wang and Han Zhou and Zaiqiao Meng},
      year={2025},
      eprint={2508.07407},
      archivePrefix={arXiv},
      primaryClass={cs.AI},
      url={https://arxiv.org/abs/2508.07407}, 
}

@misc{song2024trialerrorexplorationbasedtrajectory,
      title={Trial and Error: Exploration-Based Trajectory Optimization for LLM Agents}, 
      author={Yifan Song and Da Yin and Xiang Yue and Jie Huang and Sujian Li and Bill Yuchen Lin},
      year={2024},
      eprint={2403.02502},
      archivePrefix={arXiv},
      primaryClass={cs.CL},
      url={https://arxiv.org/abs/2403.02502}, 
}

@misc{wang2025ragenunderstandingselfevolutionllm,
      title={RAGEN: Understanding Self-Evolution in LLM Agents via Multi-Turn Reinforcement Learning}, 
      author={Zihan Wang and Kangrui Wang and Qineng Wang and Pingyue Zhang and Linjie Li and Zhengyuan Yang and Xing Jin and Kefan Yu and Minh Nhat Nguyen and Licheng Liu and Eli Gottlieb and Yiping Lu and Kyunghyun Cho and Jiajun Wu and Li Fei-Fei and Lijuan Wang and Yejin Choi and Manling Li},
      year={2025},
      eprint={2504.20073},
      archivePrefix={arXiv},
      primaryClass={cs.LG},
      url={https://arxiv.org/abs/2504.20073}, 
}
\bibliographystyle{iclr2026_conference}

\appendix

\section{Implementation Details}
\label{app:implementation_details_all_envs}
    
    \subsection{Models and Decoding}
    All results in the main paper use \texttt{gpt-5.4-mini} as the underlying LLM for every role. Decoding settings differ by role:
    \begin{itemize}
        \item \emph{Actor agent:} \texttt{reasoning\_effort = low}, \texttt{parallel\_tool\_calls = false}, request limit $100$ per episode.
        \item \emph{Inducer agent:} \texttt{reasoning\_effort = medium}, \texttt{parallel\_tool\_calls = false}, request limit $100$ per episode.
    \end{itemize}
    Libraries are reset to empty at the start of training for each task family; only the primitives carry over. During the sleep cycle, the inducer is allowed to add, edit, or remove library entries between sleep cycles.

    \subsection{Training and Evaluation Schedule}
    Each training run consists of $200$ online rollouts. The inducer is invoked every $10$ rollouts (one sleep cycle per $10$ training episodes), and the resulting library is evaluated on the $30$-episode test set every $50$ training rollouts, giving five evaluation checkpoints per run (zero-shot at rollout $0$, plus rollouts $50, 100, 150, 200$). Each setting is repeated over $3$ random seeds. At each checkpoint, we first compute each seed-level metric over the $30$ test episodes, then report the mean and $\pm 1,\text{s.d.}$ across the three seed-level values.

    \subsection{Cost Accounting}
    \label{app:cost-accounting}
Per-episode inference cost is computed as
$$
\mathrm{cost}
=
p_{\mathrm{in}} N_{\mathrm{in}}^{\mathrm{uncached}}
+
p_{\mathrm{cache}} N_{\mathrm{in}}^{\mathrm{cached}}
+
p_{\mathrm{out}} N_{\mathrm{out}},
$$
where $N_{\mathrm{in}}^{\mathrm{uncached}}$, $N_{\mathrm{in}}^{\mathrm{cached}}$, and $N_{\mathrm{out}}$ are the uncached input, cached input, and output tokens summed over all LLM calls in the episode. This includes actor calls during rollout and sleep-time inducer calls amortized over the rollouts in the corresponding inducer batch. 
We use the published API prices as of May~6,~2026. For \texttt{gpt-5.4-mini}, we set
\[
p_{\mathrm{in}} = \$0.75,\qquad
p_{\mathrm{cache}} = \$0.075,\qquad
p_{\mathrm{out}} = \$4.50
\]
per million tokens. 

For \texttt{gemini-3-flash-preview}, we set
\[
p_{\mathrm{in}} = \$0.35,\qquad
p_{\mathrm{cache}} = \$0.0875,\qquad
p_{\mathrm{out}} = \$1.05
\]
per million tokens.

We use this dollar-cost accounting in Figure~\ref{fig:combined_performance}.

\section{ScienceWorld Details}
    \label{app:scienceworld}

    \subsection{Chosen Tasks}
    We evaluate on two ScienceWorld task families: \texttt{Electricity} (family~3) and \texttt{Classification} (family~4). Each family contains multiple task names (sub-tasks); episodes are addressed by \emph{(task\_name, variation\_idx)} pairs. We use ScienceWorld's built-in train/dev/test variation partitions (\texttt{get\_variations\_train/dev/test}), which are disjoint by construction. For each family we sample \textbf{200} train variations and \textbf{30} test variations with a fixed seed ($42$); the same train/test sets are reused across all three training seeds, so only the rollout stochasticity differs across runs.

    \subsection{Action Interface}
    The agent acts through a fixed set of 25 primitive Python functions that wrap ScienceWorld's text parser. These primitives cover navigation (\texttt{go}, \texttt{look\_around}, \texttt{look\_at}, \texttt{look\_in}), object manipulation (\texttt{pick\_up}, \texttt{put\_down}, \texttt{move}, \texttt{pour}, \texttt{dunk}), state changes (\texttt{open}, \texttt{close}, \texttt{activate}, \texttt{deactivate}, \texttt{connect}, \texttt{disconnect}, \texttt{mix}), measurement and inspection (\texttt{focus\_on}, \texttt{read}, \texttt{use}, \texttt{inventory}, \texttt{task\_description}), and auxiliary actions (\texttt{eat}, \texttt{flush}, \texttt{wait}, \texttt{disambiguate}). Each primitive takes typed string arguments (e.g.\ \texttt{connect(obj\_a, obj\_b)}); this is the interface the induced library composes against. We do not expose ScienceWorld's \texttt{get\_valid\_actions} list to the agent. In our setting, the agent's effective action space includes both primitives and any library functions available under the current method; surfacing only the primitive-level valid-action set would be misleading, since it omits the higher-level abstractions the agent is expected to prefer.

    \subsection{Observations and Episode Termination}
    Observations are plain text: on each step the agent receives the room description and the parser's acknowledgement of the previous action. The full observation history is kept in context (no sliding window). The initial prompt contains the task description returned by ScienceWorld's \texttt{taskdescription()} and the opening room observation; no oracle hints or gold trajectories are provided.
    
    An episode ends when (i) the actor agent calls the \texttt{submit\_answer} tool to declare the task complete, or (ii) the actor agent exceeds its budget of \textbf{100} LLM calls per episode. We set this budget at $100$ because additional tool calls no longer meaningfully improve the agent's success rate, so further extending the budget would inflate cost without changing the headline results. In both cases, success is determined by checking ScienceWorld's internal score: the episode is scored as a success iff the score has reached its maximum value of $1.0$, and as a failure otherwise.
    
\section{BabyAI Details}
\label{app:babyai}

\subsection{Chosen Tasks}
We evaluate on the \texttt{pick\_up\_seq\_go\_to} subtask of BabyAI-Text as exposed by BALROG. Episodes are addressed by environment seed. We sample \textbf{200} train seeds and \textbf{30} test seeds with a fixed top-level seed ($42$); train and test seed sets are disjoint by construction. The same train/test sets are reused across all three training seeds, so only the rollout stochasticity differs across runs.

\subsection{Action Interface}
The agent acts through a fixed set of 6 primitive Python functions that wrap BabyAI-Text's command interface: navigation (\texttt{turn\_left}, \texttt{turn\_right}, \texttt{go\_forward}), object manipulation (\texttt{pick\_up}, \texttt{drop}), and state change (\texttt{toggle}). Each primitive takes no arguments and operates on the cell directly in front of the agent; this is the interface the induced library composes against. We do not expose BALROG's valid-action list to the agent. In our setting, the agent's effective action space includes both primitives and any library functions available under the current method; surfacing only the primitive-level valid-action set would be misleading, since it omits the higher-level abstractions the agent is expected to prefer.

\subsection{Observations and Episode Termination}
Observations are plain text descriptions of the agent's egocentric view, as produced by the BabyAI-Text wrapper. The full observation history is kept in context (no sliding window). The initial prompt contains the synthetic natural-language instruction sampled by BabyAI's Baby Language grammar and the opening observation.

An episode ends when (i) the simulator emits a terminal signal indicating the instruction has been satisfied, or (ii) the actor agent exceeds its budget of \textbf{30} LLM calls per episode. We set this budget at $30$ because additional tool calls no longer meaningfully improve the agent's success rate, so further extending the budget would inflate cost without changing the headline results. Like ScienceWorld, success is determined entirely by the environment's binary task-completion signal: the episode is scored as a success if the instruction is fully satisfied within the budget, and as a failure otherwise.

\section{Crafter Details}
\label{app:crafter}

\subsection{Episode Sampling}
Each Crafter episode is a freshly generated world keyed by a single integer seed. 
We use three independent runs, each with \textbf{200} training worlds and \textbf{30} test worlds, but with different top-level seeds to ensure diversity.
All methods are run three times using the same training and test worlds, so comparisons are matched by world seed.
Unless stated otherwise, reported means and uncertainty bands are computed over these three matched runs.

\subsection{Action Interface}
The agent acts through a fixed set of 18 primitive Python functions wrapping the Crafter engine: 17 are game-advancing primitives and one is a non-advancing read action. These primitives cover navigation (\texttt{move\_north}, \texttt{move\_south}, \texttt{move\_east}, \texttt{move\_west}), interaction (\texttt{do} -- a multi-use action that collects material, drinks from a lake, or hits the creature in front; \texttt{sleep}; \texttt{noop} -- advance one tick without acting), structure placement (\texttt{place\_stone}, \texttt{place\_table}, \texttt{place\_furnace}, \texttt{place\_plant}), and tool crafting (\texttt{make\_wood\_pickaxe}, \texttt{make\_stone\_pickaxe}, \texttt{make\_iron\_pickaxe}, \texttt{make\_wood\_sword}, \texttt{make\_stone\_sword}, \texttt{make\_iron\_sword}). All 17 game-advancing primitives are zero-argument and return \texttt{None}; each advances the Crafter world by exactly one tick. The \texttt{get\_current\_observation()} primitive returns the current textual observation without advancing the world, and is provided so that library functions can re-inspect the world between primitive calls. Crafter also has no \texttt{submit\_answer} action; episodes terminate only through the environment itself (see below). 

\subsection{Observations}
Observations are plain text rendered from the agent's egocentric $9\times9$ local view, with three blocks shown each step: vital status, inventory contents, and visible objects (with the object directly in front of the agent reported separately). Each observation is wrapped between \texttt{<<OBSERVATION\_BEGIN>>} and \texttt{<<OBSERVATION\_END>>} delimiters and substituted once into the \texttt{\{obs\}} slot of the user prompt template at the start of the rollout. It will also be returned upon agent calling the \texttt{get\_current\_observation()} primitive.

A representative mid-episode observation looks as follows:
\begin{lstlisting}
<<OBSERVATION_BEGIN>>
Your status:
- health: 7/9
- food: 5/9
- drink: 6/9
- energy: 8/9
Your inventory:
- wood: 3
- stone: 1
- wood_pickaxe: 1
You see:
- grass 1 step north
- tree 2 steps north and 1 step east
- stone 3 steps east
- zombie 4 steps north and 2 steps west
- water 5 steps south and 3 steps east
You face grass at your front.
<<OBSERVATION_END>>
\end{lstlisting}
When the agent's inventory is empty, the inventory block collapses to the line ``\emph{You have nothing in your inventory.}''; when nothing other than the player is visible in the $9\times9$ window, the visible-objects block collapses to ``\emph{You see nothing away from you.}'' Sleeping or dead states are surfaced by prefixing the status block with a single sentence (e.g., ``\emph{You are sleeping, and will not be able to take actions until energy is full.}'').

\subsection{Episode Termination}
An episode ends when (i) the player's health reaches zero, or (ii) the agent exhausts its budget of \textbf{2{,}000} primitive actions. Each episode is scored as the fraction of the 22 achievements unlocked at termination. The actor agent's per-episode LLM-call budget (\texttt{request\_limit}) is also set to $2{,}000$ to ensure that code baselines are not disadvantaged by being cut off early.. In practice, we rarely see any rollout exceeding this budget.

\subsection{Achievements}
The agent is told about the 22 Crafter achievements in the initial system prompt, in a fixed display order: Collect Wood, Place Table, Eat Cow, Collect Sapling, Collect Drink, Make Wood Pickaxe, Make Wood Sword, Place Plant, Defeat Zombie, Collect Stone, Place Stone, Eat Plant, Defeat Skeleton, Make Stone Pickaxe, Make Stone Sword, Wake Up, Place Furnace, Collect Coal, Collect Iron, Make Iron Pickaxe, Make Iron Sword, Collect Diamond.

\subsection{Zombie-Frequency Variants}
\label{app:crafter-zombie}
The external-randomness analysis in \S\ref{sec:analysis:external-randomness} sweeps a single environment knob, \texttt{zombie\_frequency} ($f$), while holding all other settings fixed. The knob enters the upstream Crafter engine in two places. First, at world reset: when $f = 0$, every zombie generated by terrain initialization is explicitly removed before the first step. Second, at every tick during world balancing: zombies spawn with probability $\min(1.0,\ 0.3 f)$ and despawn with probability $\min(1.0,\ 0.4 f)$ per eligible chunk, and the per-chunk target population is scaled by $f$. With this scaling, $f = 1.0$ recovers the upstream Crafter defaults, $f = 0.0$ removes zombies entirely, and $f = 2.0$ doubles both spawn and despawn pressures (each clipped at $1.0$) and doubles the target population. No other engine parameters are changed across variants.

For the external randomness experiments, we instantiate this knob at three levels labeled \textbf{0x} ($f=0.0$), \textbf{1x} ($f=1.0$, default), and \textbf{2x} ($f=2.0$), and run each of the three comparison methods at each level: OPO, Voyager, and \method. For a given method, the 1x configuration is the same as the one used in the main results; therefore we directly reused the results from the main experiments; the 0x and 2x configurations differ only in \texttt{zombie\_frequency}.

\section{ASI Adaptation Details}
\label{app:asi_adapt}

We adapt ASI \citep{wang2025inducingprogrammaticskillsagentic} to our \emph{purly}-online setting while preserving its programmatic skill induction mechanism. In the original algorithm, each newly induced skill undergoes a \emph{skill verification} step, in which the agent attempts to re-solve the inducing task using the new skill and retains the skill only if verification succeeds. Because this step requires additional task interaction beyond the original online trajectory, we remove it to satisfy our no-replay constraint. All other components of ASI are left unchanged.

\section{Voyager Adaptation Details}
\label{app:voyager_adapt}

Given its relevance to our work, we make three task-level adaptations to preserve Voyager's \citep{wang_voyager_2023} core ingredients while aligning its training protocol with our online constraints. 
First, the curriculum agent is informed about the Crafter's 22 achievements as the final optimization target when proposing the next intermediate subgoals. 
Second, we remove Voyager's respawn-style continuation after death: when the agent dies or the episode terminates, training restarts from a fresh Crafter world with no inventory carryover, matching the irreversible online setting used by all other methods. Third, evaluation is performed through the same rollout interface used by the other baselines, so that learned skills are assessed under the same acting protocol rather than through a bespoke execution regime.

\section{Additional Main Results}

\label{app:additional_results}
\subsection{Code Interpreter Ablations}
\label{app:code_interpreter_ablation}

\begin{figure*}[t]
    \centering
    \includegraphics[width=\textwidth]{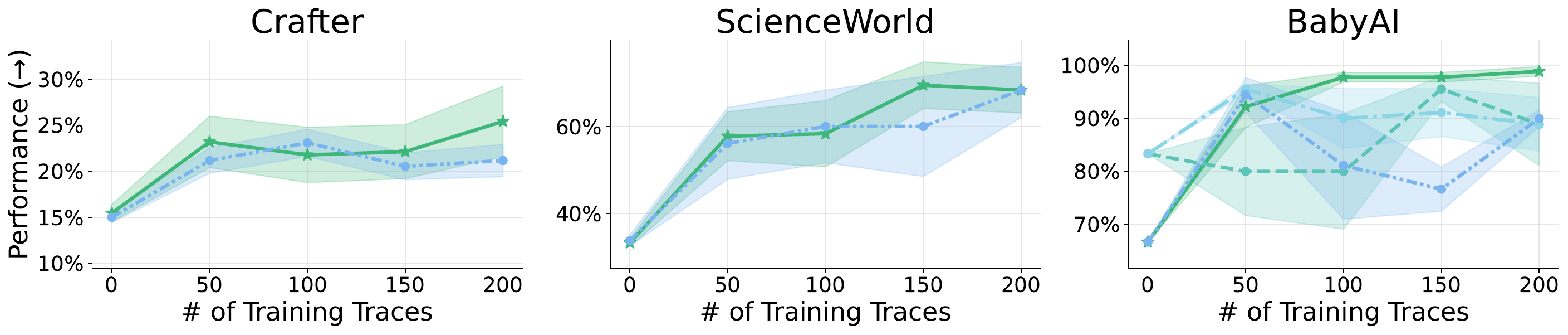}
    \hfill
    \includegraphics[width=\textwidth]{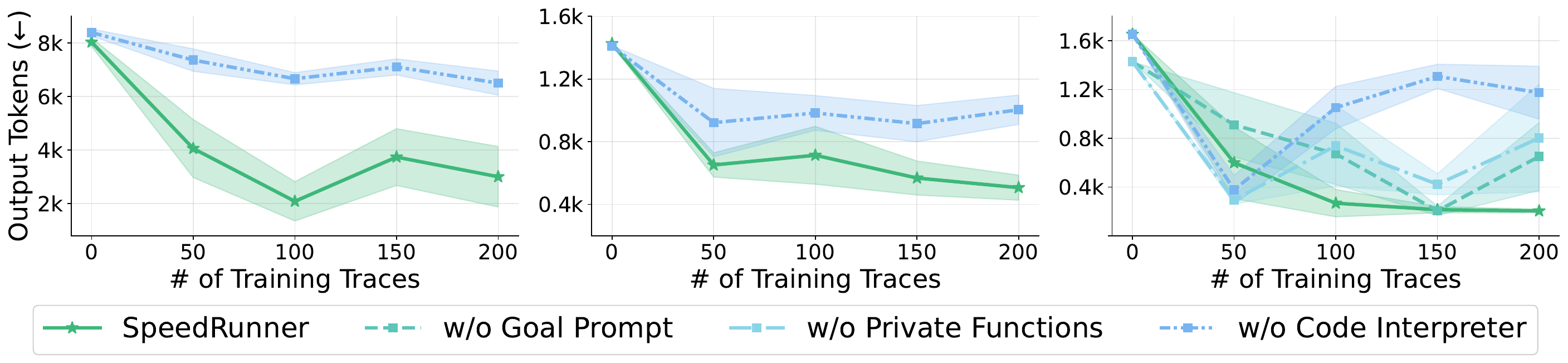}
    \caption{Extra \method ablations on the Crafter and ScienceWorld environments. As in BabyAI, removing access to the code interpreter has only small effects on performance but contributes substantially to improvements in efficiency.}
    \label{fig:combined_ablations}
\end{figure*}

Our results in Figure \ref{fig:combined_ablations} show that removing the code interpreter degrades final performance in two out of three environments and increases the amount of output tokens required substantially in all three. This demonstrates the importance of this component for \method's gains in performance and especially efficiency for programmatic skill induction.
 
  \subsection{Cross-Model Results}
  \label{app:gemini_results}

\begin{figure*}[t!]
    \centering
    \includegraphics[width=\textwidth]{figures/combined_baselines_perf_gpt-5.4-mini.pdf}

    \vspace{0.4em}

    \includegraphics[width=\textwidth]{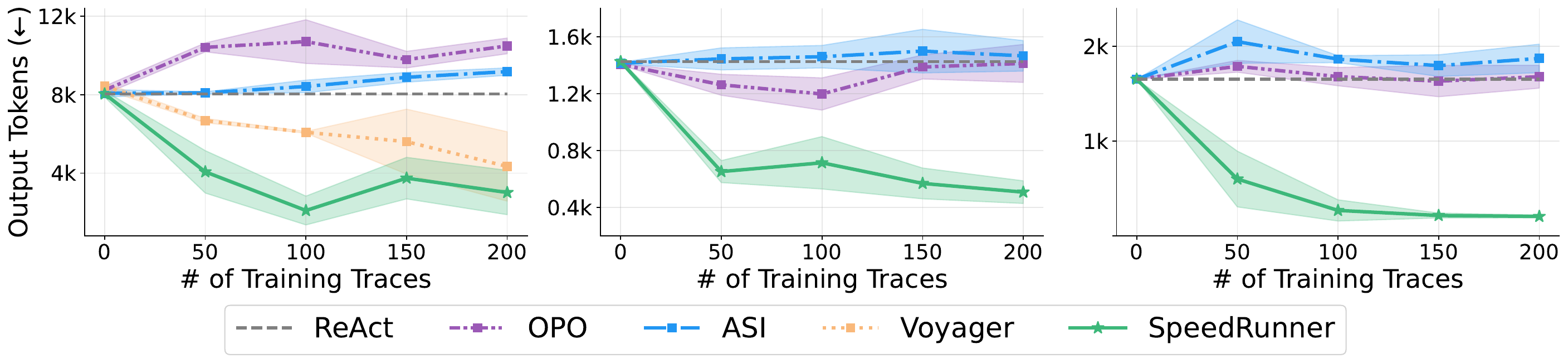}

    \caption{
    Main results using GPT-5.4-mini with tokens instead of cost for easy comparison across models. 
    }
    \label{fig:combined_performance_gpt_tokens}
\end{figure*}

\begin{figure*}[t!]
    \centering
    \includegraphics[width=\textwidth]{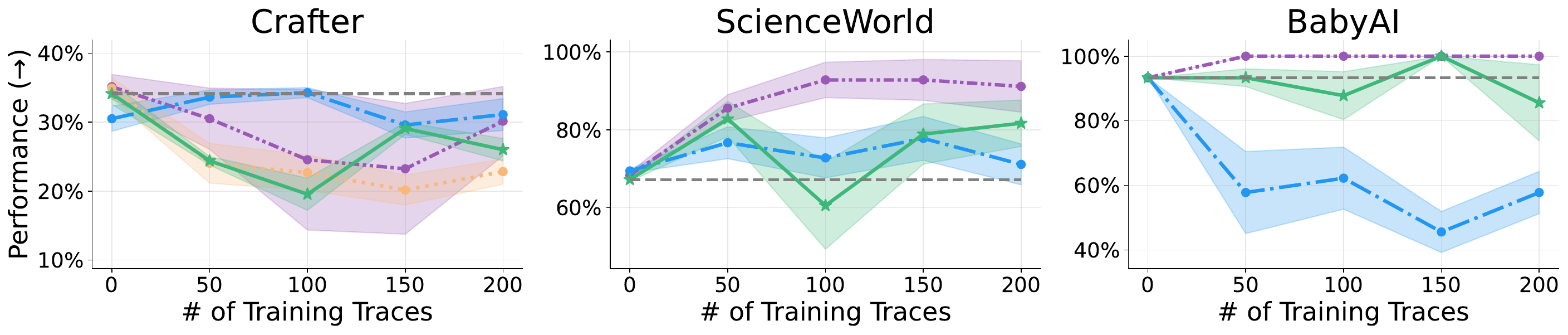}

    \vspace{0.4em}

    \includegraphics[width=\textwidth]{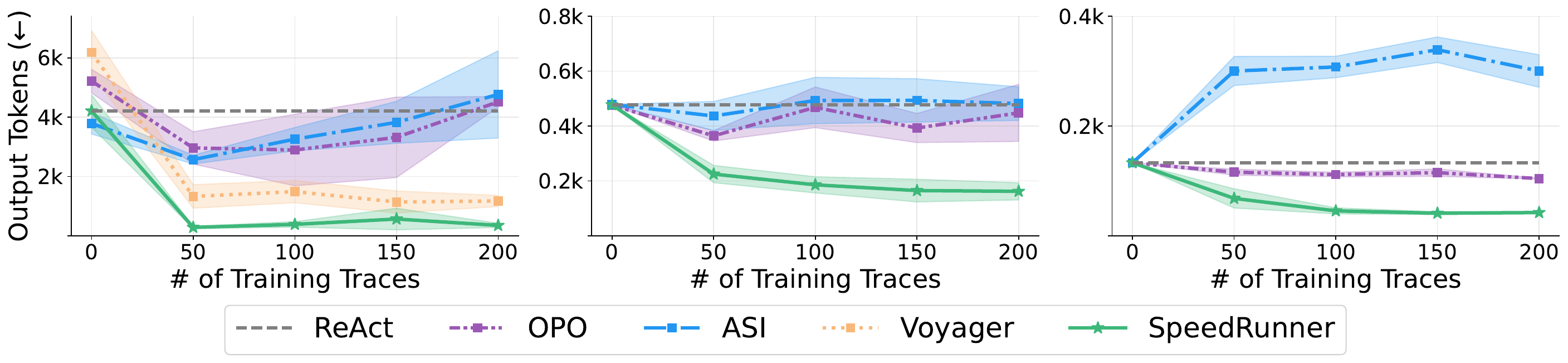}

    \caption{
    Gemini-3-Flash performance and output-token cost across Crafter, ScienceWorld, and BabyAI.
    As with GPT-5.4-mini, \method reduces output tokens across the board, however, performance improvements are much less consistent than with GPT-5.4-mini.}
    \label{fig:combined_performance_gemini}
\end{figure*}

\begin{figure*}[t!]
    \centering
    \includegraphics[width=\textwidth]{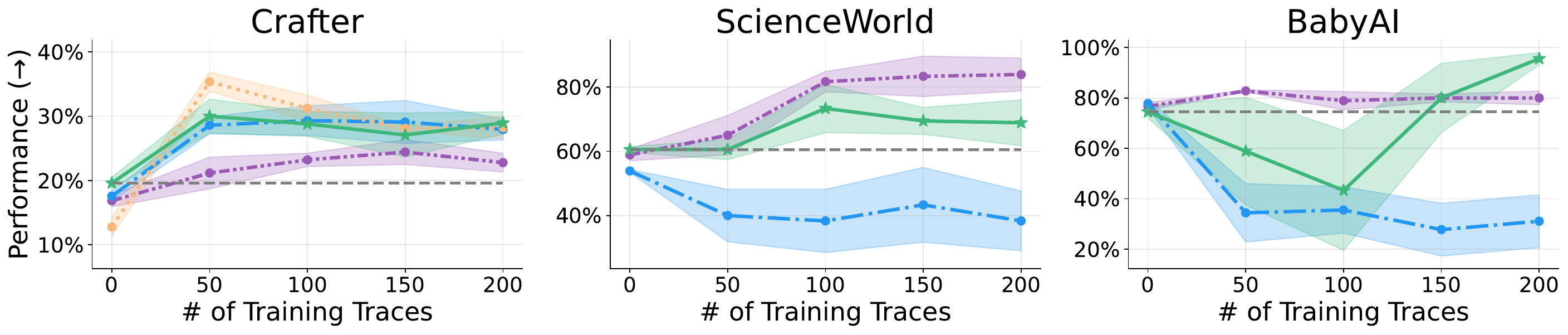}

    \vspace{0.4em}

    \includegraphics[width=\textwidth]{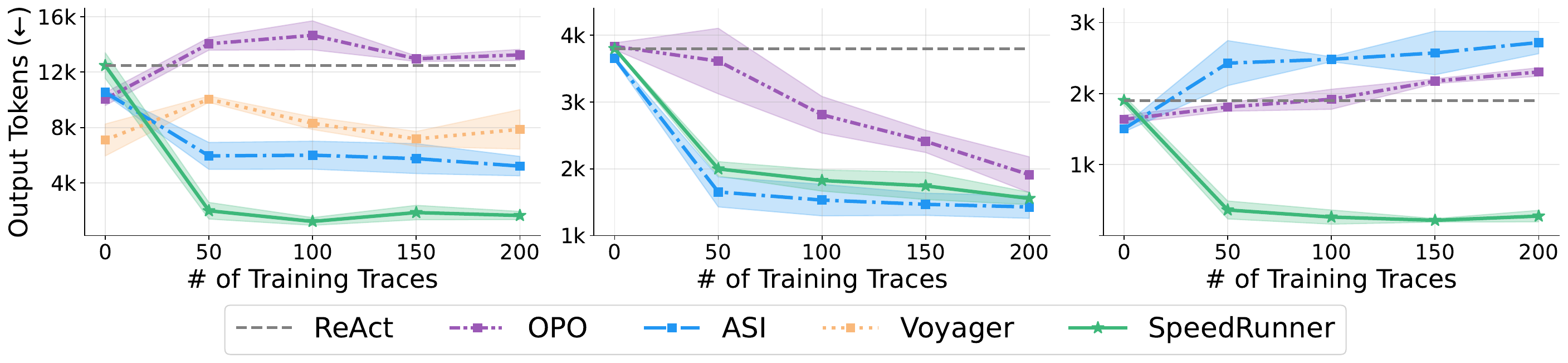}

    \caption{
    Qwen-3.5-27B performance and output-token cost across Crafter, ScienceWorld, and BabyAI.
    As with GPT-5.4-mini and Gemini-3-Flash, \method generally matches or improves final task performance while reducing output tokens. 
    }
    \label{fig:combined_performance_qwen_tokens}
\end{figure*}

In order to test whether \method's trends depend on the backbone LLM, we repeat the main comparison with Gemini-3-Flash and Qwen-3.5-27B. As shown in Figures~\ref{fig:combined_performance_gemini} and \ref{fig:combined_performance_qwen_tokens}, \method retains the best output-token efficiency across the board and the best performance--efficiency tradeoff in most instances. More specifically, using the same two-sided pair t-test from our main results, we find that our method is significantly more efficient and better performing than our code-based baselines in all settings with every models except for Crafter, where they are much more competitive. Additionally, our method's performance against OPO, our non-code baseline, reveals an interesting pattern. \method achieves significantly worse final success rate than OPO with Qwen in ScienceWorld and Gemini in all benchmarks, despite achieving much lower token usage. We highlight this result because it shows that aggressive code delegation is not always performance-improving when the base actor is already strong. Not all performance can be transferred into code.

In an attempt to understand this behavior, we note that both Gemini-3-Flash and Qwen-3.5-27B start from higher performance levels than GPT-5.4-mini in all benchmarks where OPO beats \method, indicating stronger raw model capacity in these environments. 
We believe this behavior is driven by compression aggressiveness.
\method’s iterative induction produces hierarchical libraries that delegate a large fraction of decisions to code; in Crafter, for example, Gemini-3-Flash reduces LLM calls by 94\% over training. 
This yields very low inference cost, but it can also make the policy too rigid in stochastic environment. 
Code induced from earlier trajectories may encode procedures that are efficient on average but less able to adapt to later situations requiring reactive judgment. 
Thus, these results expose a real tradeoff: \method can over-compress a capable actor, improving cost while sacrificing some performance. 

\begin{figure*}[t!]
    \centering
    \includegraphics[width=\textwidth]{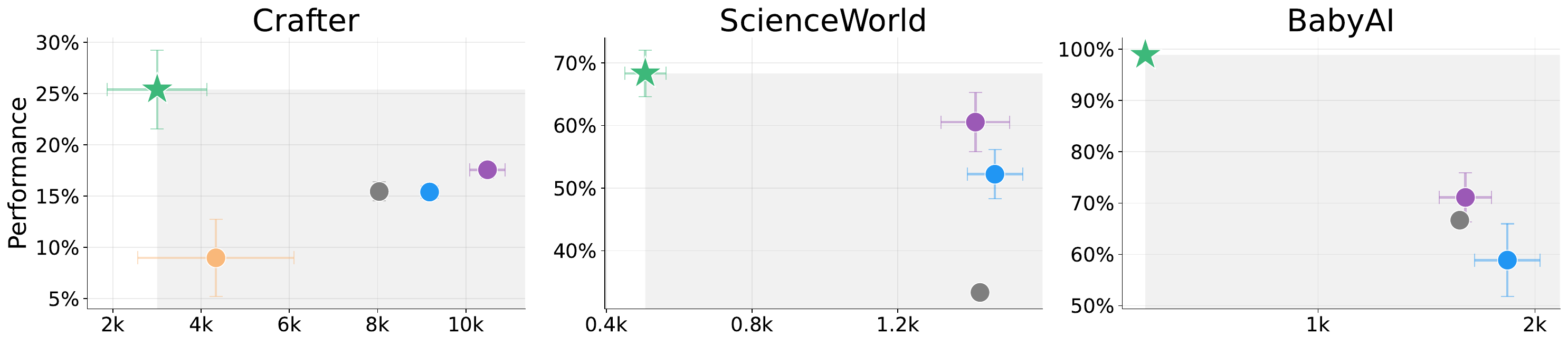}

    \vspace{0.4em}

    \includegraphics[width=\textwidth]{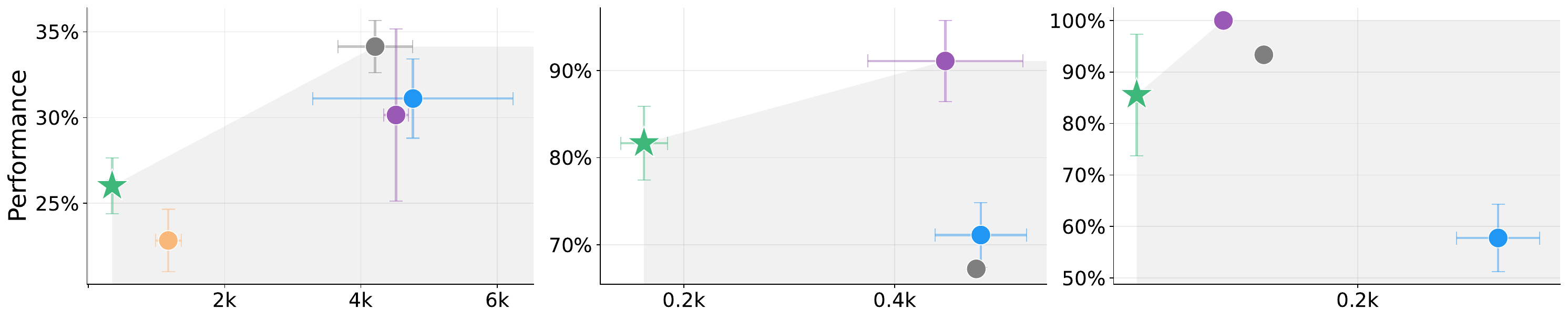}

    \vspace{0.4em}

    \includegraphics[width=\textwidth]{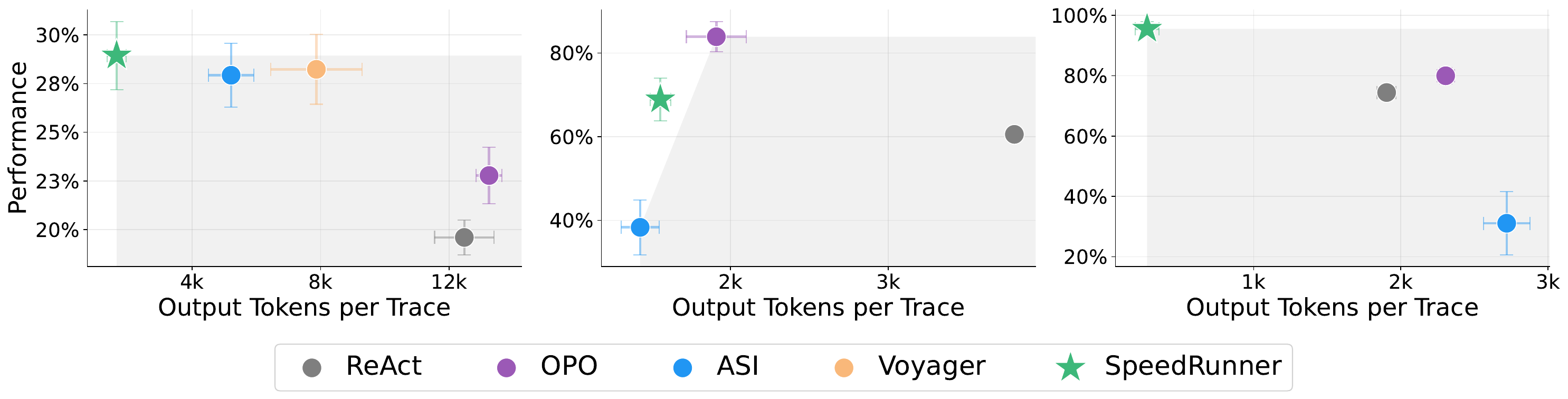}

    \caption{
     Performance vs output tokens per trace for all approaches and benchmarks for GPT-5.4-mini (top), Gemini-3-Flash (middle) and Qwen-3.5-27B (bottom) for easy comparison. This figure demonstrates that SpeedRunner dominates in most settings or remains at the frontier in all experiments we conducted.
    }
    \label{fig:pareto_frontier}
\end{figure*}

Nevertheless, as illustrated in Figure \ref{fig:pareto_frontier}, our approach succeeds in \emph{consistent} learning and efficiency performance.
Of the 9 settings, our approach dominates in 5 and is on the frontier in the remainder, a feat that is unmatched by any of the other baselines. 

 \subsection{Offline Continual Learning with ASI}
\label{app:replay_ablation}
\begin{figure}[t]
    \centering
    \includegraphics[width=0.8\textwidth]{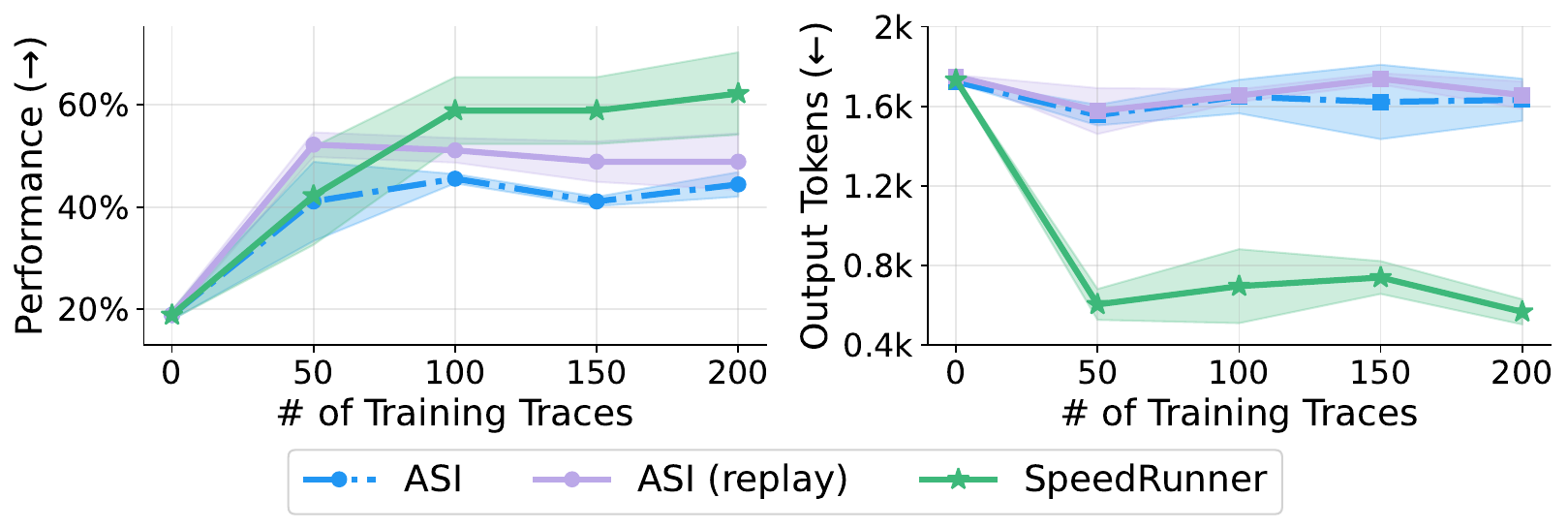}
    \caption{Effect of ASI's replay-based skill verification on ScienceWorld task~3. Restoring replay-based verification improves ASI's task performance, particularly early in training.\method achieves the highest final performance while using substantially fewer output tokens than both ASI variants.}
    \label{fig:replay_ablation}
\end{figure}

We further study the effect of experience replay by comparing our online adaptation and the original offline version of ASI. As described in Appendix \ref{app:asi_adapt} , the original algorithm verifies each new skill by re-executing the inducing task, whereas our online adaptation removes this step.

Figure \ref{fig:replay_ablation} compare ASI, ASI (replay), and \method on ScienceWorld task~3. Restoring the verification step improves ASI's mean task performance, particularly early in training,because it filters out broken skills that lead to regression on a previous task. However, it does not reduce token cost beyond the adapted online version. After 200 training traces, \method achieves higher mean task performance than both ASI variants while using substantially fewer output tokens per episode (approximately 65\% fewer than both ASI variants). These results indicate that removing ASI's replay-based verification step makes continual skill induction more challenging.

\subsection{Why Performance Gains Vary by Benchmark}
\label{app:per-benchmark-gains}

\method's main advantage comes from converting repeated behavior into compact executable routines, and the value of this conversion depends on the structure of the environment.  ScienceWorld and BabyAI contain many reusable procedures that can be invoked repeatedly once discovered, so \method\ improves performance while dramatically compressing output length. The effect is sharpest on BabyAI, where final per-eval-epoch output reaches roughly an \emph{eighth} of the ReAct baseline --- approaching an order-of-magnitude compression ratio that, combined with near-perfect final accuracy, indicates that the actor is almost entirely delegating decisions to a small set of induced abstractions rather than re-deriving primitive sequences.  BabyAI's high zero-shot baseline ($\sim$$67\%$) limits how much headroom remains for raw performance gains, but this compression is itself the evidence that the code delegation pays off in reliability \emph{and} efficiency.

Crafter is more challenging: the environment is longer-horizon, stochastic, and reactive, so a routine that is useful in one state can become inappropriate when immediate survival needs arise (e.g.\ a zombie attack).  This shrinks the room for raw performance gains, but the compression advantage persists --- \method's end-of-training output tokens are more than $3\times$ lower than OPO's and ASI's, even though all four methods begin from the same ReAct cost baseline and ASI and OPO actually grow rather than shrink.  We analyze the structural reorganization that lets \method\ retain this advantage as environmental randomness rises in \S\ref{sec:analysis:external-randomness}.

\section{Full Quantitative Codebook Analysis}
\label{app:quantitative}

We provide the full structural analysis of the learned codebooks here. The main text summarizes the two most diagnostic call-graph measures, maximum depth and density; this appendix reports the complete set of statistics, including library size, average depth, and cyclomatic complexity.

Recall that we characterize the codebooks learned by each method using five structural measures, computed from the final codebook at the end of training. We first construct a directed call graph $G=(V,E)$ for each codebook, where each node $v \in V$ is an induced function and each edge $(u,v) \in E$ indicates that function $u$ calls function $v$. Let $n=|V|$.

\begin{itemize}[leftmargin=*, itemsep=2pt, topsep=2pt]
  \item \textbf{Library size} (\textit{\#Fns}) is the number of induced functions, including both public and private functions.
  \item \textbf{Maximum depth} (\textit{max d.}) is the length of the longest directed path in $G$, measuring the height of the deepest abstraction built by the inducer.
  \item \textbf{Average depth} (\textit{avg d.}) is the mean depth across all induced functions in $G$.
  \item \textbf{Density} is $|E|/(n(n-1))$, excluding self-edges, and measures how often induced functions call other induced functions rather than relying only on the primitive action set.
  \item \textbf{Cyclomatic complexity} (\textit{cyclo.}) is the standard McCabe complexity averaged across functions, capturing whether individual skills encode fixed action sequences or branch over multiple cases.
\end{itemize}

\begin{table*}[t!]
\centering
\small
\setlength{\tabcolsep}{4pt}
\begin{tabular}{llccccc}
\toprule
\textbf{Benchmark} & \textbf{Method} & \textbf{\#Fns} & \textbf{Max d.} & \textbf{Avg d.} & \textbf{Density} & \textbf{Cyclo.} \\
\midrule
\multirow{4}{*}{Crafter}
  & ASI & $33.7${\sd{7.2}} & $1.7${\sd{0.6}} & $0.61${\sd{0.06}} & $0.030${\sd{0.008}} & $2.4${\sd{1.1}} \\
  & Voyager & $1{,}077${\sd{565}} & $5.7${\sd{2.1}} & $0.44${\sd{0.21}} & $0.0005${\sd{0.0003}} & $\bm{30.2}${\sd{7.2}} \\
  & \method\ w/o CI & $23.3${\sd{11.5}} & $6.3${\sd{3.2}} & $2.27${\sd{1.33}} & $0.138${\sd{0.078}} & $10.9${\sd{5.1}} \\
\rowcolor{oursrow}
  & \textbf{\method\ (ours)} & $\bm{22.3}${\sd{4.0}} & $\bm{8.7}${\sd{2.1}} & $\bm{2.78}${\sd{0.96}} & $\bm{0.149}${\sd{0.042}} & $17.5${\sd{3.2}} \\
\addlinespace
\multirow{3}{*}{ScienceWorld (task 4)}
  & ASI & $56.7${\sd{42.1}} & $1.3${\sd{0.6}} & $0.61${\sd{0.26}} & $0.015${\sd{0.006}} & $1.4${\sd{0.6}} \\
  & \method\ w/o CI & $\bm{25.3}${\sd{4.5}} & $5.0${\sd{1.0}} & $1.38${\sd{0.34}} & $\bm{0.072}${\sd{0.007}} & $\bm{11.7}${\sd{4.7}} \\
\rowcolor{oursrow}
  & \textbf{\method\ (ours)} & $30.0${\sd{3.0}} & $\bm{6.0}${\sd{1.0}} & $\bm{1.99}${\sd{0.63}} & $0.064${\sd{0.015}} & $9.1${\sd{1.1}} \\
\addlinespace
\multirow{3}{*}{ScienceWorld (task 3)}
  & ASI & $73.0${\sd{12.2}} & $2.0${\sd{0.0}} & $0.84${\sd{0.27}} & $0.017${\sd{0.002}} & $1.0${\sd{0.1}} \\
  & \method\ w/o CI & $31.3${\sd{6.7}} & $5.3${\sd{1.5}} & $1.63${\sd{0.39}} & $\bm{0.074}${\sd{0.018}} & $\bm{9.1}${\sd{2.5}} \\
\rowcolor{oursrow}
  & \textbf{\method\ (ours)} & $\bm{30.7}${\sd{1.5}} & $\bm{5.7}${\sd{0.6}} & $\bm{1.84}${\sd{0.12}} & $0.069${\sd{0.020}} & $8.5${\sd{2.1}} \\
\addlinespace
\multirow{3}{*}{BabyAI}
  & ASI & $25.3${\sd{6.7}} & $1.0${\sd{0.0}} & $0.41${\sd{0.12}} & $0.018${\sd{0.005}} & $2.0${\sd{1.6}} \\
  & \method\ w/o CI & $9.0${\sd{4.0}} & $\bm{3.3}${\sd{1.5}} & $\bm{1.31}${\sd{0.70}} & $0.178${\sd{0.019}} & $8.1${\sd{1.1}} \\
\rowcolor{oursrow}
  & \textbf{\method\ (ours)} & $\bm{5.7}${\sd{0.6}} & $2.3${\sd{0.6}} & $0.73${\sd{0.40}} & $\bm{0.206}${\sd{0.042}} & $\bm{10.9}${\sd{1.7}} \\
\bottomrule
\end{tabular}
\caption{Structural measures of each method's final skill library across four benchmark settings, using GPT-5.4-mini with mean$\pm$sd over three seeds. Shaded rows mark the full \method\ system. Bold marks the most compact library for \#Fns and the highest value for the remaining structural measures within each benchmark.}
\label{tab:complexity_global}
\end{table*}
Table~\ref{tab:complexity_global} shows that \method\ learns compact but substantially more structured libraries than the baselines. Relative to ASI, \method\ consistently increases call-graph density, maximum depth, and average depth, indicating that its skills are not merely shallow wrappers around primitive actions but reusable routines with hierarchical dependencies. Relative to Voyager on Crafter, \method\ achieves much higher density with nearly two orders of magnitude fewer functions, suggesting that it factors repeated behavior into shared helpers instead of accumulating scenario-specific code.

Although Voyager has higher cyclomatic complexity on Crafter, this complexity is concentrated inside individual functions rather than distributed through reusable library structure. Its density is only $0.0005$, compared with $0.149$ for \method, meaning that its functions rarely call one another. Thus, Voyager's library is large and internally branchy, but weakly compositional: each function tends to encode a complete scenario rather than decomposing behavior into smaller reusable components.

The code-interpreter ablation preserves much of \method's organization, but the full method is generally deeper, denser, or more decision-rich. On Crafter and both ScienceWorld settings, \method\ has higher maximum and average depth than \method\ w/o CI. On BabyAI, where the tasks are shorter and the learned libraries are naturally small, \method\ instead has the highest density and cyclomatic complexity. Overall, \method's advantage is not larger codebooks, but better organized ones: compact libraries whose skills are more reusable, modular, and compositional.

\section{Per-Environment Qualitative Analysis}
\label{app:qualitative}

This appendix expands the three patterns identified in the main paper with concrete examples from each environment. Each subsection contrasts the key abstraction discovered by ASI, the code-interpreter-free ablation (\method w/o CI), and \method.

\subsection{BabyAI: Compositional Mission Parsing}
\label{sec:abstractions:babyai}

The PickUpSeqGoTo task requires parsing a natural-language mission (e.g., \emph{``go to the red box after you pick up the blue ball''}) and executing it in a partially-observable grid world. Some phrasings reverse grammatical order relative to execution order, which is the decisive challenge.

\begin{table*}[ht]
\centering
\caption{Key abstractions per method in the BabyAI environment. The task hinges on parsing missions where grammatical order can reverse execution order. ASI templates motion sequences from traces and never parses the mission; \method w/o CI patches navigation failures reactively but still defers mission parsing to the LLM; only \method derives an explicit priority-ordered parser by executing candidate patterns against historical mission strings.}
\label{tab:babyai}
\small
\resizebox{\textwidth}{!}{\begin{tabular}{@{}l p{0.85\linewidth}@{}}
\toprule
\textbf{Method} & \textbf{Key abstraction} \\
\midrule
ASI & Fixed motion sequences (\texttt{repeat\_left\_step\_and\_look}, \texttt{move\_left\_then\_pick\_up}) extracted verbatim from successful traces. No function reads observations or branches on state. \\
\addlinespace
\method w/o CI & \texttt{\_approach\_object} encodes three failure modes (target hidden, blocked, laterally offset), each guard added one sleep phase after the failure that revealed it. No mission parser. \\
\addlinespace
\method & \texttt{\_parse\_mission\_goals}: priority-ordered pattern list where reversed-order phrasings are matched before the left-to-right fallback. Discovered by running candidate parsers against historical mission strings via \texttt{execute\_code}. \\
\bottomrule
\end{tabular}}
\end{table*}

A naïve left-to-right parser inverts execution order on reversed-phrasing missions. \method w/o CI has no parser and leaves disambiguation to the LLM, which occasionally misorders goals (its 6.7\% error). \method's coding agent iterated over historical mission strings inside \texttt{execute\_code}, observed the inversion, and added reordering entries to a priority-ordered pattern list (matching \emph{``go to X after you pick up Y''} and \emph{``pick up X after you go to Y''} before the fallback). The gap between \method w/o CI and \method is entirely attributable to this one structural decision.

\subsection{Crafter}
\label{sec:abstractions:crafter}

\begin{table*}[ht]
\centering
\caption{Crafter: key abstractions per method, showing how each handles the tech-tree progression and resource-lookup failures characteristic of the environment. ASI crafts unconditionally with no inventory checks; Voyager appends a new function per episode and never consolidates, yielding 483 near-duplicate variants; \method w/o CI builds reactive guards one failure at a time; \method encodes the full progression hierarchy in a single function and diagnoses systematic lookup failures via trace queries to introduce \texttt{\_bounded\_exploration}. The Voyager row illustrates that observation-reading alone is insufficient---consolidation is a necessary co-condition.}
\label{tab:crafter}
\small
\resizebox{\textwidth}{!}{\begin{tabular}{@{}l p{0.85\linewidth}@{}}
\toprule
\textbf{Method} & \textbf{Key abstraction} \\
\midrule
ASI & \texttt{prepare\_basic\_wood\_tools} crafts unconditionally regardless of inventory; observation calls at function ends are trace artefacts. \\
\addlinespace
\method w/o CI & \texttt{bootstrap\_stone\_pickaxe} gates stone behind a wood-pickaxe check; hostile-check before rest in \texttt{stabilize\_visible\_survival} added reactively after deaths. Guards accumulate one failure at a time. \\
\addlinespace
\method & \texttt{\_choose\_progress\_target} encodes the full tech-tree hierarchy in one function, enumerated via \texttt{execute\_code} before any episode. \texttt{\_bounded\_exploration} added after the inducer diagnosed a high rate of failed resource lookups across batches. \\
\addlinespace
Voyager & New function appended per episode; no editing. 212 near-identical \texttt{collect\_one\_wood\_log\_*} variants. Same 14-line observation-parsing block copy-pasted into every function. \\
\bottomrule
\end{tabular}}
\end{table*}

\begin{table*}[ht]
\centering
\caption{ScienceWorld T3 (conductivity): key abstractions per method. The task requires wiring an unknown substance into a battery-and-bulb circuit and placing it in the correct box based on whether the bulb lights. The decisive ASI--ablation gap is programmatic result reading: ASI places the substance after a fixed wait regardless of bulb state, while \method w/o CI gates placement on a regex over the bulb observation. \method's additional contribution is consolidation---hoisting per-function ambiguity handling into a single wrapper, cutting library size by 22\%.}
\label{tab:sw_t3}
\small
\resizebox{\textwidth}{!}{\begin{tabular}{@{}l p{0.85\linewidth}@{}}
\toprule
\textbf{Method} & \textbf{Key abstraction} \\
\midrule
ASI & Polarity functions \texttt{setup\_conductivity\_test} and \texttt{finish\_conductivity\_check} added sequentially after wiring failures. Neither clears prior circuit state; placement is unconditional after a fixed \texttt{wait()}. \\
\addlinespace
\method w/o CI & \texttt{\_test\_conductivity\_with\_circuit} reads the bulb via \texttt{re.search("which is on")}; \texttt{\_reset\_circuit} fully disconnects between attempts. Four polarity patterns added reactively. \\
\addlinespace
\method & Same programmatic result reading as \method w/o CI, plus a single \texttt{\_call\_one\_arg\_action\_with\_disambiguation} wrapper that routes every primitive call through one ambiguity handler. Room search uses DFS up to 24 steps---discovered via \texttt{execute\_code} for cases where the target is not in the stated room. \\
\bottomrule
\end{tabular}}
\end{table*}

\method's \texttt{\_choose\_progress\_target} encodes the full tech tree in one place: higher-tier ores are gated behind the corresponding pickaxe tier, and stone is de-prioritized once the stockpile is sufficient. Neither ASI nor \method w/o CI has an analogue---ASI crafts unconditionally, and \method w/o CI's guards are reactive and incomplete.

The second decisive abstraction, \texttt{\_bounded\_exploration}, illustrates the inducer's diagnostic loop. After observing repeated \texttt{not\_found} returns from \texttt{travel\_to\_visible\_resource}, the inducer queried the last batch of traces and printed every top-level \texttt{not\_found} alongside its target (wood, tree, cow). It then walked the trace tree to count call frequencies per batch and observed that \texttt{travel\_to\_visible\_resource} was being invoked 116--250 times per batch in later rounds---i.e., the agent was repeatedly trying and failing to find resources in view. Diagnosing this as a visibility problem rather than a navigation bug, the inducer added \texttt{\_bounded\_exploration}, which sweeps the local neighborhood around the agent for the desired resource before falling back to a \texttt{not\_found}.

Voyager is the counterfactual to consolidation: 483 functions, no shared progression model, boilerplate duplicated hundreds of times, and the lowest score of any method (14.7\%, below ASI's 20.5\%). Reading observations is not sufficient; consolidation is a necessary co-condition.

\subsection{ScienceWorld: Programmatic Verification}
\label{sec:abstractions:scienceworld}

ScienceWorld T3 (conductivity) requires wiring an unknown substance into a battery-and-bulb circuit and placing it in the correct box. T4 (find category) requires locating an object of a given category across rooms.

The decisive gap between ASI and \method w/o CI is programmatic result reading: ASI moves to the target box after a fixed wait regardless of bulb state; \method w/o CI gates placement on \texttt{re.search("which is on")} of the bulb observation. \method's additional contribution is consolidation---in \method w/o CI, ambiguity handling was added ad-hoc inside each function as the error appeared; \method hoisted it into a single wrapper, cutting library size 22\% (27 $\to$ 21) while applying the fix uniformly.

\begin{table*}[ht]
\centering
\caption{ScienceWorld T4 (find category): key abstractions per method, the one setting where \method w/o CI outperforms \method. Both methods program category classification---a clear advance over ASI's full delegation to the LLM---but with different strategies: \method w/o CI maintains targeted exclusion lists for observed false positives (``painting,'' ``egg''), while \method derives general normalization rules via \texttt{execute\_code}. The result suggests pre-commit testing helps most when the failure-mode space is combinatorially large; for finite enumerable false-positive sets, reactive per-failure refinement is sufficient.}
\label{tab:sw_t4}
\small
\resizebox{\textwidth}{!}{\begin{tabular}{@{}l p{0.85\linewidth}@{}}
\toprule
\textbf{Method} & \textbf{Key abstraction} \\
\midrule
ASI & \texttt{focus\_roundtrip\_pick\_move}: round-trip navigation template from traces. Category classification fully delegated to the LLM. \\
\addlinespace
\method w/o CI & \texttt{\_extract\_animal\_name} with priority-ordered patterns and artwork/egg exclusions, derived from false-positive failures (``painting of a bee''). \texttt{\_try\_pickup} verifies retrievability before returning a match. \\
\addlinespace
\method & \texttt{\_canonicalize\_object\_name} strips stage labels (``in the dead stage'') via \texttt{execute\_code}-tested rules. \texttt{\_probe\_bee\_hive\_for\_target} added after observing missed hive contents. \\
\bottomrule
\end{tabular}}
\end{table*}

T4 is the one setting where \method w/o CI outperforms \method. Both implement programmatic category classification---a clear advance over ASI's delegation---but with different strategies: \method w/o CI uses targeted exclusion lists (skip ``painting,'' ``egg''), while \method uses general normalization rules. That \method w/o CI still wins suggests pre-commit testing helps most when the failure-mode space is \emph{combinatorially large}---as in BabyAI's mission ordering or Crafter's tech tree---rather than a finite enumerable set of false-positive patterns, where reactive per-failure refinement is sufficient.

\section{External Randomness: Additional Analysis}
\label{app:zombie_library_table}

Table~\ref{tab:zombie_library_methods} reports the full structural measures (definition in \S\ref{app:quantitative}) that the figure abstracts over.

The mechanism behind \method's \%Combat rise is concretely visible in
the 2x library: of its $\sim 17$ combat-aware functions, eleven are
\emph{existing} harvest and movement routines (e.g.,
\texttt{harvest\_visible}, \texttt{clear\_front\_blocker},
\texttt{opening\_turn}) that have been edited to accept an
\texttt{include\_combat} or \texttt{allow\_combat} keyword argument ---
one kwarg flip and every caller gets the defensive variant.  Voyager,
unable to edit, instead spawns a fresh bespoke function for each new
encounter (\texttt{collect\_wood\_log\_avoid\_front\_zombie},
\texttt{collect\_one\_log\_avoiding\_east\_zombie},
\texttt{step\_adjacent\_to\_skeleton}, \dots): the same combat logic
that \method\ amortizes once across its call graph is paid in
duplicated branches inside every Voyager skill, which is why Voyager's
per-skill cyclomatic remains the highest of any method even as its
density stays near zero.

\begin{table*}[t!]
\centering
\small
\setlength{\tabcolsep}{4pt}
\begin{tabular}{llcccccc}
\toprule
\textbf{Method} & \textbf{Cond.} & \textbf{\#Fns} & \textbf{Max d.}
                & \textbf{Avg d.} & \textbf{Density} & \textbf{Cyclo.}
                & \textbf{\%Combat} \\
\midrule
Voyager   & 0x & $1{,}368$ & $5$       & $0.40$ & $0.0002$ & $33.4$ & $0.7\%$  \\
Voyager   & 1x & $1{,}077$ & $5.7$     & $0.44$ & $0.0005$ & $30.2$ & $2.9\%$  \\
Voyager   & 2x & $590$     & $9$       & $0.52$ & $0.0007$ & $31.6$ & $11.5\%$ \\
\addlinespace
\method   & 0x & $25$      & $6$       & $2.40$ & $0.132$  & $11.2$ & $24.0\%$ \\
\method   & 1x & $22$      & $8.7$     & $2.78$ & $0.149$  & $17.5$ & $31.5\%$ \\
\method   & 2x & $32$      & $\bm{10}$ & $2.62$ & $0.094$  & $14.8$ & $\bm{53.1\%}$ \\
\bottomrule
\end{tabular}
\caption{Final-library structure across the three randomness conditions (GPT-5.4-mini). Columns are the structural measures from \S\ref{app:quantitative}; \emph{\%Combat} is the fraction of induced functions whose body references a creature/combat keyword.} \label{tab:zombie_library_methods}
\end{table*}

\section{Post-hoc Library Truncation on Voyager}
\label{app:voyager_truncation}

A natural reading of Voyager's regression under standard ($1\times$) and doubled ($2\times$) zombie pressure (Table~\ref{tab:zombie_library_methods}) is that its append-only library simply grows past the point where the rollout agent can usefully attend to it: by the end of training, the GPT-5.4-mini seeds carry libraries of $\{615, 909, 1707\}$ public skills, and one of them already overflows the model's context window outright. If volume alone were the issue, then keeping only the most recently induced skills should recover most of the gap to the no-codebook baseline. We therefore ask the following question: can a simple last-$N$ truncation of Voyager's final library rescue its $1\times$-zombie performance?

\paragraph{Setup.}
We take the final codebook checkpoint of one $1\times$-zombie Voyager seed and re-evaluate it on the same $30$-episode Crafter test set used in the main results (\S\ref{sec:main-results}), holding the rollout agent and seeds fixed. For each $N \in \{50, 100, 200, 400, 800\}$, we present the rollout agent with only the \emph{last} $N$ skills---those induced most recently---as callable tools and as entries in the instruction manual; the earlier $909-N$ skills are hidden from the prompt but remain in the execution namespace, so any visible skill whose body invokes an earlier one still runs correctly. This isolates the variable of interest: the size of the \emph{agent-visible} library, with the underlying behavior of each visible skill held fixed. We bracket the sweep with two reference points reused from the training-time periodic evaluation of the same run: $N=0$ corresponds to the no-codebook ReAct baseline (the actor runs with primitives only), and $N=909$ corresponds to the full final library.

\begin{table*}[t]
\centering
\small
\setlength{\tabcolsep}{6pt}
\begin{tabular}{l ccccccc}
\toprule
\textbf{Visible $N$}  & 0 (none) & 50 & 100 & 200 & 400 & 800 & 909 (full) \\
\midrule
\textbf{Mean Progression}     & 0.173    & 0.124 & 0.129 & 0.130 & \textbf{0.174} & 0.127 & 0.153 \\
\bottomrule
\end{tabular}
\caption{Post-hoc truncation of Voyager's final $1\times$-zombie library ($909$ induced skills). The rollout agent is shown only the most recently induced $N$ skills; earlier skills remain callable when invoked from inside a visible one. Mean Progression is the fraction of the $22$ Crafter achievements unlocked per episode, averaged over the $30$-episode held-out test set. $N=0$ and $N=909$ are the no-codebook and full-library reference points taken from the training-time periodic evaluation of the same run.}
\label{tab:voyager_truncation}
\end{table*}

\paragraph{Truncation does not structurally remedy Voyager.}
Three observations from Table~\ref{tab:voyager_truncation} point against volume being the binding constraint. First, the truncation curve is \emph{non-monotonic} in $N$: there is no clear rising or falling trend as the visible library shrinks. Second, \emph{no truncation meaningfully exceeds the no-codebook baseline}: the closest setting, $N=400$, matches it within noise (a gap of one achievement unlocked on one episode), and every other setting falls below. No choice of $N$---including the smallest visible surface we test---turns Voyager's library into a net asset relative to running with primitives only. Third, the full library also underperforms the no-codebook baseline on this seed, consistent with the regression pattern reported in Table~\ref{tab:zombie_library_methods}.

Taken together, these results indicate that Voyager's regression under standard zombie density is not a pure context-bloat phenomenon; the \emph{composition} of the induced library matters more than the headcount of skills exposed at rollout time.

    \section{Use of AI Assistants}
We used ChatGPT and Claude as assistants for writing (grammar correction, clarity edits, and suggestions on presentation) and coding (implementation and debugging of experimental infrastructure). All scientific ideas, experimental designs, analyses, claims, and final text were reviewed and approved by the authors.

\end{document}